\documentclass[10pt,twocolumn,letterpaper]{article}

\usepackage[pagenumbers]{cvpr} % To force page numbers, e.g. for an arXiv version

\usepackage{xspace}
\usepackage{multirow}
\usepackage{array}
\usepackage{tikz}
\usetikzlibrary{arrows.meta}
\newcommand{\paper}{EgoGeneralist\xspace}

\AtEndPreamble{\creflabelformat{equation}{#2#1#3}}

\definecolor{cvprblue}{rgb}{0.21,0.49,0.74}
\usepackage[pagebackref,breaklinks,colorlinks,allcolors=cvprblue]{hyperref}

\title{{EgoGeneralist: Unifying Egocentric Action--Video Tasks \\through Video-Native Modeling}}

\author{
{\fontsize{11}{14}\selectfont
Yanlin Jin$^{1,2}$ \hspace{1.5em}
Bowen Yang$^{2}$ \hspace{1.5em}
Cheng Zhang$^{1}$ \hspace{1.5em}
Yifan Yang$^{3,2}$
}\\[2pt]
{\fontsize{11}{14}\selectfont
Zhe Li$^{2}$ \hspace{2em}
Jianfei Cai$^{1,\dagger}$ \hspace{2em}
Kai Zhu$^{2,\dagger}$
}\\[5pt]
{\small
$^{1}$Monash University \quad
$^{2}$Ant Digital Technologies, Ant Group \quad
$^{3}$Institute of Automation, Chinese Academy of Sciences
}\\[-2pt]
{\small
$^{\dagger}$Corresponding authors
\qquad
Project page:
\href{https://www.neiljin.site/egogeneralist_page/}
{\texttt{egogeneralist\_page}}
}
}

\makeatletter
\g@addto@macro\@maketitle{% !TeX root = ../main.tex
\begin{center}
  \begin{minipage}{\textwidth}
    \centering
    \includegraphics[width=\textwidth]{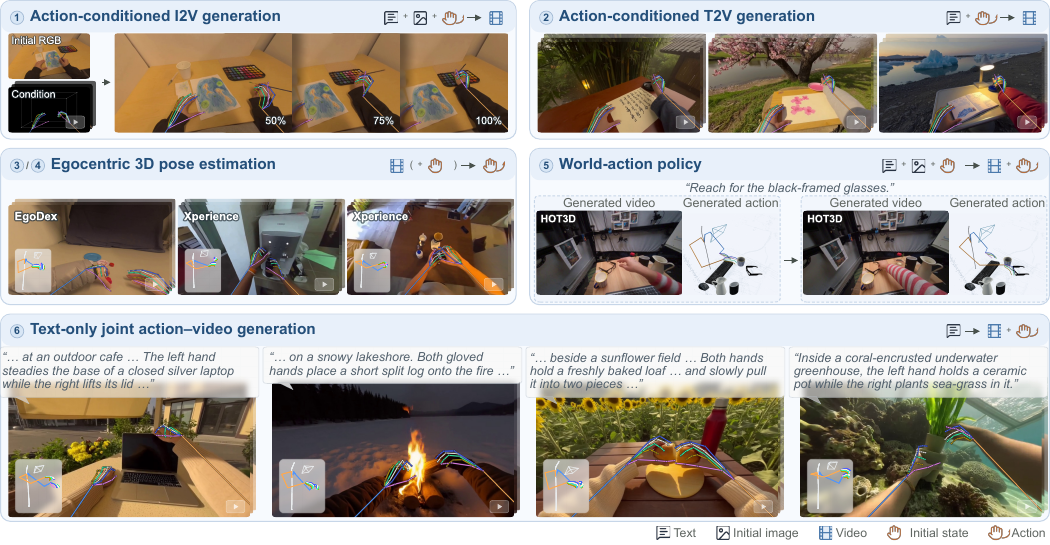}
    \captionof{figure}{\textbf{\paper} unifies six action--video tasks across three prediction directions: action-conditioned ego video generation, egocentric 3D pose estimation, and joint action--video generation.
    Each direction supports inference with or without an initial target state.
    Reusing the pretrained video prior enables generalization to unseen scenes with physically coherent interactions.}
    \label{fig:teaser-preview}
  \end{minipage}
\end{center}
}
\makeatother

\makeatletter
\patchcmd{\@maketitle}
  {\vspace*{12pt}}
  {\vspace*{8pt}}
  {}
  {\PackageWarning{title-layout}{Bottom spacing patch failed}}

\patchcmd{\@maketitle}
  {\vspace*{24pt}}
  {\vspace*{12pt}}
  {}
  {\PackageWarning{title-layout}{Title-author spacing patch failed}}
\makeatother

\begin{document}
\maketitle
% !TeX root = ../main.tex
\begin{abstract}
Modern video models possess strong visual priors, yet do not natively represent human action in metric 3D space.
We present \paper, which adapts pretrained video generators to unify action-conditioned video generation, pose estimation, and joint action--video generation for egocentric manipulation.
Our role-adaptive action renderer provides video-native representations for multidirectional action--video learning, making metric motion usable as both a spatial control signal and a recoverable generation target.
Cross-stream action--video coupling establishes close correspondence between metric motion and visual dynamics through lightweight adaptation while preserving the pretrained visual prior.
Together, these designs establish a generalist framework spanning six action--video tasks.
With only narrow-domain data, \paper generalizes to diverse unseen scenes and interactions with effective action control and visual quality, bridging pretrained visual knowledge and metric 3D human motion.
\par
\end{abstract}

% !TeX root = ../main.tex

\section{Introduction}
\label{sec:introduction}

Modern video generators have acquired strong priors over appearance, motion, and visual interactions~\cite{wan2025wan}.
These priors make them a great foundation for egocentric video generation, which is central to applications in embodied learning, human--object interaction, and AR/VR~\cite{luo2025beingh0,grauman2022ego4d,xie2026generatedreality}.
Yet in egocentric manipulation, what a person sees is jointly shaped by body motion, hand articulation, and camera movement.
A video generator can synthesize such experiences, but it does not natively represent what the person does in metric 3D space. Bridging embodied motion and its visual consequences is therefore essential.

This connection is inherently multidirectional.
Actions can drive video generation; conversely, video observations provide evidence for recovering motion; and joint generation captures how action and appearance evolve together (\cref{fig:teaser-preview}).
Although typically studied in isolation, these problems are simply different directions of the same action--video relationship.
Recent unified systems cover multiple directions, but still rely on separately parameterized action representations~\cite{zhu2025uwm,li2025uva,sun2026handworld,wang2026donk}.
Aligning such representations with video often involves large-scale joint training, limiting reuse of the generative prior~\cite{wang2026donk,bi2026motus,nvidia2026cosmos3}.
This raises a central question: \textit{Can a pretrained video generator become an egocentric action--video generalist, extending its visual prior to metric 3D motion and generalizing broadly?}

\begin{figure*}[t]
    \centering
    \includegraphics[width=0.99\textwidth,trim=6bp 0bp 6bp 0bp,clip]{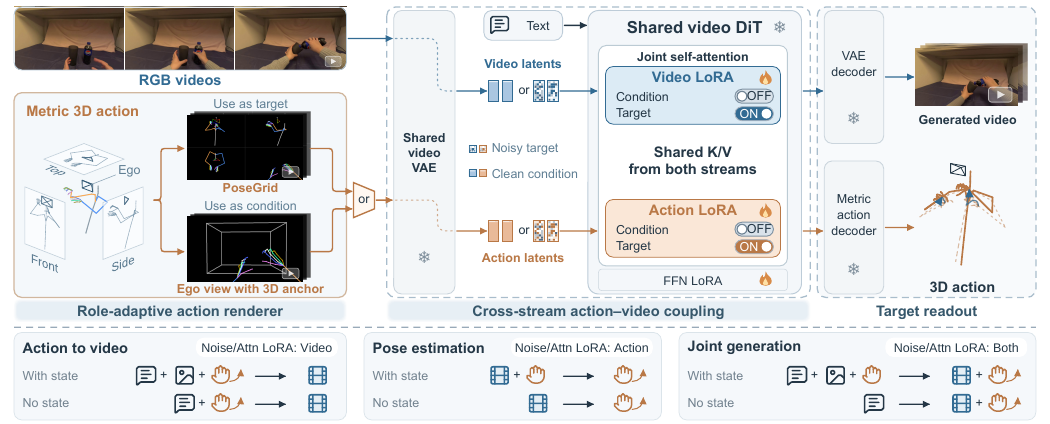}
    \captionsetup{width=\textwidth}
    \caption{\textbf{Overview of \paper.} Role-adaptive rendering (\cref{sec:action-interfaces}) tailors the visual representation of metric 3D action to its task role: an ego-view map with a fixed 3D anchor for conditioning, or a multiview Posegrid for generation and metric readout.
    A shared frozen video VAE encodes rendered action and video into two latent streams.
    Cross-stream coupling (\cref{sec:action-video-learning}) uses joint self-attention and attention LoRAs activated only for generation targets, retaining the backbone weights and per-stream input structure.
    The bottom panel illustrates condition--target assignments for the three prediction directions defined in \cref{sec:problem-formulation}.}
    \label{fig:action-video-denoising}
\end{figure*}

To this end, we introduce \paper, a video-native framework that unifies action--video learning in egocentric manipulation (\cref{fig:action-video-denoising}).
A \emph{role-adaptive renderer} represents action as video, which the pretrained video model can natively consume as a condition or generate as a target.
\emph{Cross-stream action--video coupling} then establishes correspondence through shared attention and target-specific attention LoRA in the video latent space, without extra input channels or conditioning encoders.

Our framework supports six action--video tasks (\cref{fig:teaser-preview}).
These capabilities encompass action-conditioned video synthesis, action recovery from video, and joint action--video generation with metric upper-body, hand, and camera motion.
Despite being trained only on controlled laboratory recordings, the adapted video generators generalize to unseen scenes, objects, and interactions.
From text alone, they further synthesize paired metric action and video, offering a new means of action--video data generation.
Our contributions are as follows:
\begin{itemize}

\item \textbf{A unified framework for egocentric action--video learning.} We formulate action-conditioned video synthesis, metric action recovery, and joint action--video generation as a single video-based problem, coupling visual generation with metric upper-body, hand, and camera motion within a unified framework.
 
\item \textbf{Role-adaptive rendering with cross-stream action--video coupling.} We introduce role-adaptive action renderings for spatial control and metric action recovery, coupled with video within the pretrained model's native latent space through shared attention and low-rank adaptation.
This design learns action--video correspondence with less disruption to the visual prior.

\item \textbf{Unified capabilities competitive with task-specific experts.} Experiments across multiple tasks demonstrate that our framework, adapted only on narrow-domain data, matches or surpasses specialized baselines, while generalizing to in-the-wild action--video generation.
\end{itemize}

% !TeX root = ../main.tex

\section{Related Work}
\label{sec:related_work}

\subsection{Action-Conditioned Ego Video Generation}
A central design choice in action-conditioned egocentric video generation is how motion is represented and fused with the denoising network.
Embedding-based methods encode action sequences and inject them into the video model, e.g., through adaptive normalization~\cite{bai2025peva}, timestep embeddings~\cite{bagchi2026egowm}, or feature modulation and cross-attention~\cite{pallotta2026egocontrol}, with camera motion folded into the same representation~\cite{pallotta2026egocontrol} or encoded as camera-ray features~\cite{xie2026generatedreality}.
Some methods further augment action embeddings with encoded skeleton renderings~\cite{xie2026generatedreality} or occlusion-aware reference features propagated along hand trajectories~\cite{zhang2026controllable}.
Across these designs, motion control still relies on aligning dedicated motion embeddings with the pretrained video features.

Spatially aligned methods instead render prescribed motion in the image plane, using projected skeletons~\cite{wang2025vap,gao2026lome,wu2026oscar,hao2026egosim} or hand-mesh renderings~\cite{kim2026dwm} as explicit spatial guidance.
These renderings, however, are typically coupled to the generator through expanded input channels~\cite{kim2026dwm,hao2026egosim} or auxiliary conditioning branches~\cite{wang2025vap}, which alter the pretrained computation path and can weaken reuse of the visual prior~\cite{long2024wonder3d,kocsis2025intrinsix,chen2026unividx}.
Camera control poses a further challenge, since body and hand projections leave viewpoint motion implicit.
Existing approaches embed camera motion~\cite{gao2026lome,xie2026generatedreality} or use reconstructed scenes and 3D scene memory for viewpoint conditioning~\cite{kim2026dwm,hao2026egosim,yan2026egogenesis}, the latter requiring additional reconstruction and scene maintenance.
In contrast, our framework provides explicit action and camera conditioning without scene reconstruction or separate camera embeddings.
Moreover, rendered motion in these methods serves only as a condition, whereas ours supports both spatial control and metric action readout within the native latent space of the pretrained video model.

\subsection{Egocentric 3D Pose Estimation}
Egocentric 3D pose estimation seeks to recover body and hand motion from first-person observations.
Hand reconstruction has advanced from single-image estimation~\cite{park2022handoccnet,pavlakos2024hamer,potamias2025wilor} to video-based methods that exploit temporal cues for pose estimation~\cite{wen2023htt} and account for camera motion to recover world-space hand trajectories~\cite{zhang2025hawor,yu2025dynhamr}.
Extending reconstruction to the body further requires inferring motion from incomplete and distorted observations~\cite{tome2019xregopose}.
Prior work addresses this ambiguity through geometric context and learned motion priors.
Specifically, external-view supervision during training, scene geometry, and stereo-temporal observations provide constraints on body configuration~\cite{wang2022egopw,wang2023sceneego,akada2024egostereo}.
Learned motion priors also support estimation conditioned on head or camera trajectories~\cite{li2023egoego,yi2025egoallo,patel2025uniegomotion}.
Modeling correlations between body and hand motion further enables their joint recovery~\cite{yi2025egoallo,wang2024egowholemocap,lee2025rewind}.
More recent work jointly reconstructs human motion, camera trajectories, and scene geometry from monocular egocentric video~\cite{lyu2026reviv}.
Unlike these methods, our framework reformulates upper-body, hand, and camera motion recovery as a unified video generation task, jointly recovering metric 3D motion through a pretrained video denoiser.

\subsection{Unified Egocentric Action--Video Learning}
Unified multimodal diffusion supports multiple prediction directions within a shared model by varying which modalities are observed and generated~\cite{bao2023unidiffuser,le2025onediffusion,chen2026unividx}.
In robotics, this principle has been used to couple video prediction with action inference~\cite{zhu2025uwm,li2025uva,bi2026motus}.
Closely related works extend unified modeling to egocentric interactions, jointly generating video and hand actions~\cite{sun2026handworld,wang2026donk}.
However, they only consider hand motion and rely on action-specific encoders, denoisers, or conditioning networks.
In contrast, our framework jointly models metric upper-body, hand, and camera motion and couples action and video through lightweight adaptation.
By retaining the pretrained backbone and per-stream input structure, we effectively reuse the visual prior and achieve strong generalization.

% !TeX root = ../main.tex

\section{Method}
\label{sec:method}

\subsection{Problem Formulation}
\label{sec:problem-formulation}
Our goal is to enable pretrained video generators to synthesize egocentric video under action control, to recover action from video, and to jointly generate action and video.
Let $\mathbf{V}=\{\mathbf{v}_t\}_{t=1}^{F}$ denote an egocentric manipulation video and $\mathbf{A}=\{\mathbf{a}_t\}_{t=1}^{F}$ its synchronized metric 3D action sequence.
Each action state comprises upper-body and hand joint poses, together with the camera pose, all expressed in the initial camera frame.
We denote a text prompt by $\mathbf{p}$.

We consider three directions, each with two tasks depending on whether an initial target state is provided.

\paragraph{Action-conditioned egocentric video generation.}
Generate video from prescribed actions and text, optionally given the first frame:
\begin{subequations}
\label{eq:action-to-video}
\refstepcounter{equation}\label{eq:action-to-video-full}%
\refstepcounter{equation}\label{eq:action-to-video-initial}%
\begin{equation*}
\text{(a)}\; (\mathbf{A},\mathbf{p})\rightarrow\mathbf{V}
\qquad
\text{(b)}\; (\mathbf{A},\mathbf{v}_1,\mathbf{p})\rightarrow\mathbf{V}_{2:F}
\tag{\theparentequation}
\end{equation*}
\end{subequations}

\paragraph{Egocentric pose estimation.}
Recover metric 3D action from video, optionally given the first action state:
\begin{subequations}
\label{eq:video-to-action}
\refstepcounter{equation}\label{eq:video-to-action-full}%
\refstepcounter{equation}\label{eq:video-to-action-initial}%
\begin{equation*}
\text{(a)}\; \mathbf{V}\rightarrow\mathbf{A}
\qquad
\text{(b)}\; (\mathbf{V},\mathbf{a}_1)\rightarrow\mathbf{A}_{2:F}
\tag{\theparentequation}
\end{equation*}
\end{subequations}

\paragraph{Joint action--video generation.}
Generate synchronized action and video from text, optionally given the initial frame and action state:
\begin{subequations}
\label{eq:joint-action-video}
\refstepcounter{equation}\label{eq:joint-action-video-full}%
\refstepcounter{equation}\label{eq:joint-action-video-initial}%
\begin{equation*}
\text{(a)}\; \mathbf{p}\!\rightarrow\!(\mathbf{V},\mathbf{A})
\qquad
\text{(b)}\; (\mathbf{p},\mathbf{v}_1,\mathbf{a}_1)\!\rightarrow\!(\mathbf{V}_{2:F},\mathbf{A}_{2:F})
\tag{\theparentequation}
\end{equation*}
\end{subequations}

In each pair, the first mapping generates the complete target sequence, while the second predicts its continuation from the provided initial state.
We next describe how action is encoded according to its role in each task (\cref{sec:action-interfaces}) and how action and video are coupled while preserving the video prior (\cref{sec:action-video-learning}).

\subsection{Role-Adaptive Action Rendering}
\label{sec:action-interfaces}
To ease action--video alignment and better leverage the pretrained visual prior, we render action as video.
Action rendering, however, serves two roles.
As a condition, its spatial alignment with the egocentric view yields more direct correspondence with video in the spatially structured latent space.
As a generation target, it must preserve 3D geometry for metric readout, which a single ego-view projection leaves ambiguous (ablations in \cref{sec:supp-rendering-ablation}).
We therefore introduce a role-adaptive action renderer that selects between two video representations according to the role of action.

\paragraph{Ego-view action map for action conditioning.}
For visual control, we project action $\mathbf{A}$ onto the egocentric image plane under the same camera and resolution as the RGB video.
The resulting spatial alignment directly indicates where body (primarily forearms) and hands should appear and move.
Semantic colors identify body parts and fingers, while perspective changes in projected size provide depth cues.
We omit explicit color coding of depth or coordinates, which we found to provide little additional benefit in practice.

The remaining ambiguity is camera motion.
A moving skeleton alone provides no static reference to separate camera motion from body and hand motion.
We address this with a sparse, fixed-size \emph{3D anchor cage} defined by two rectangular planes at depths of $0.25$ and $0.70$ m, with corresponding corners connected.
The cage remains fixed in the initial camera frame $\mathcal{C}_s$ and is reprojected at each timestep, encoding camera motion through changes in its projection.
For anchor $\mathbf{g}_j\in\mathbb{R}^3$, its projection at time $t$ is

\begin{equation}
    \begin{bmatrix}\mathbf{u}_{t,j}\\1\end{bmatrix}
    = \frac{1}{Z_{t,j}}\,\mathbf{K}\,
    T_{\mathcal{C}_t\leftarrow\mathcal{C}_s}
    \begin{bmatrix}\mathbf{g}_j\\1\end{bmatrix},
\end{equation}
where $\mathbf{K}$ is the camera intrinsic matrix and $Z_{t,j}$ is the anchor's depth in the current camera frame $\mathcal{C}_t$.
Because the anchors are static in $\mathcal{C}_s$, their image motion directly reveals camera translation and rotation.
Together, the aligned skeleton and 3D anchors form $\mathbf{A}_{\mathrm{ego}}$, the action condition used for video generation, as illustrated in \cref{fig:action-video-denoising,fig:wm-video-comparison}.

\paragraph{Multiview Posegrid for action readout.}
To explicitly preserve 3D information, we render the human action sequence $\mathbf{A}$ as \emph{Posegrid} $\mathbf{A}_{\mathrm{multi}}$, a $2\!\times\!2$ video canvas combining front, side, and top orthographic skeleton projections with the egocentric view (\cref{fig:action-video-denoising}).
All poses are expressed in the initial camera frame $\mathcal{C}_s$.
The complementary orthographic views expose depth and articulation that a single projection collapses.
A camera frustum records camera translation and rotation relative to $\mathcal{C}_s$.
Consistent colors distinguish body parts and fingers, while depth-aware rendering places nearer skeleton segments in front of farther ones at projected crossings.

By retaining explicit geometric information, \emph{Posegrid} makes metric readout readily learnable.
We encode $\mathbf{A}_{\mathrm{multi}}$ with the frozen video VAE $\mathcal{E}$ and train an action decoder $\mathcal{D}$ to recover joint and camera translations and 6D rotations:
\begin{equation}
    \widehat{\mathbf{A}}
    = \mathcal{D}\!\left(\mathcal{E}
      (\mathbf{A}_{\mathrm{multi}})\right).
\end{equation}
The decoder is easily trained independently of the video model and then frozen.
Decoder architecture and training details are provided in \cref{sec:supp-posegrid-decoder} of the supplementary material.
\emph{Posegrid} thus provides a video-native readout space that maps generated latents back to metric action.

\subsection{Cross-Stream Action--Video Coupling}
\label{sec:action-video-learning}
We process the latents of video $\mathbf{V}$ and the role-selected action rendering $\mathbf{A}_r$, $r\in\{\mathrm{ego},\mathrm{multi}\}$, as separate streams of a shared denoiser.
Each stream keeps the pretrained input structure, while attention across streams enables action--video interaction (\cref{fig:action-video-denoising}).

\paragraph{Cross-stream interaction.}
The shared frozen encoder maps these inputs to $\mathbf{z}^{v}=\mathcal{E}(\mathbf{V})$ and $\mathbf{z}^{a}=\mathcal{E}(\mathbf{A}_{r})$.
Motivated by prior cross-domain diffusion designs~\cite{long2024wonder3d,kocsis2025intrinsix}, we pack these latents along the batch dimension and process them with the same DiT.
This preserves each stream's channel width, spatiotemporal layout, and input projection.
Within self-attention, each stream retains its own queries and positional encoding but attends to keys and values from both streams:
\begin{equation}
    \mathbf{H}_{m}
    =\operatorname{Attn}\!\left(
      \mathbf{Q}_{m},
      [\mathbf{K}_{v};\mathbf{K}_{a}],
      [\mathbf{V}_{v};\mathbf{V}_{a}]
    \right), \qquad m\in\{v,a\}.
\end{equation}
The shared context allows action and video to exchange information without a separate fusion network.

\paragraph{Unified action--video denoising.}
Building on this latent interaction, each task $\tau$ specifies the prediction direction by assigning streams to observed conditions $\mathcal{C}_{\tau}$ or generation targets $\mathcal{T}_{\tau}$ (\cref{fig:action-video-denoising}).
Action conditions use the ego-view map, whereas action targets use \emph{Posegrid} (\cref{sec:action-interfaces}).

We use separate self-attention LoRA adapters~\cite{hu2022lora} for action and video to learn modality-specific denoising from the shared action--video context.
To reduce interference with condition processing, we adopt target-gated attention adaptation~\cite{chen2026unividx}, activating adapters only for generation targets.
For each projection and modality $m\in\{v,a\}$,
\begin{equation}
    \mathbf{h}_{m}
    = W_{0}\mathbf{x}_{m}
    + g_{\tau,m}B_{m}A_{m}\mathbf{x}_{m},
    \qquad
    g_{\tau,m}=\mathbb{I}[m\in\mathcal{T}_{\tau}],
\end{equation}
where $\mathbf{x}_m$ is the projection input, $W_0$ is the frozen pretrained weight, and $A_m,B_m$ are modality-specific low-rank matrices.
The binary task gate $g_{\tau,m}$ leaves condition-stream attention projections unchanged.
We also apply a shared FFN LoRA to adapt token features in both streams~\cite{chen2026unividx}.

We then train all three mappings with one conditional flow-matching objective~\cite{lipman2023flow}, following unified multimodal diffusion~\cite{le2025onediffusion,chen2026unividx}.
Initial observations and action states use the backbone's native first-frame conditioning.
We perturb each latent stream as
\begin{equation}
    \widetilde{\mathbf{z}}^{m}_{t}
    = (1-\sigma_m)\mathbf{z}^{m}
    + \sigma_m\boldsymbol{\epsilon}^{m},
\end{equation}
where $\boldsymbol{\epsilon}^{m}\sim\mathcal{N}(\mathbf{0},\mathbf{I})$ is Gaussian noise.
The noise level $\sigma_m$ is sampled for target streams and fixed at the minimum training noise level for condition streams.
Let $\widetilde{\mathbf{Z}}_{t}$ collect the perturbed action and video streams.
The denoiser predicts $\mathbf{v}^{m}_{\theta}(\widetilde{\mathbf{Z}}_{t},t,\mathbf{p}_{\tau})$, and we minimize the mean squared velocity error over the target streams:
\begin{equation}
    \mathcal{L}_{\tau}
    =\mathbb{E}\!\left[
      \frac{1}{|\mathcal{T}_{\tau}|}
      \sum_{m\in\mathcal{T}_{\tau}}
      \operatorname{MSE}\!\left(
      \mathbf{v}^{m}_{\theta},\,
      \boldsymbol{\epsilon}^{m}-\mathbf{z}^{m}
      \right)
    \right].
\end{equation}
For \emph{Posegrid} targets, we double the squared-error contribution of latent locations covering upper-limb, hand, or camera geometry.
The text condition $\mathbf{p}_{\tau}$ is an empty prompt when text is not used.
Changing only which streams act as conditions and which are denoised yields the three mappings without task-specific denoisers.

% !TeX root = ../main.tex

\section{Experiments}
\label{sec:experiments}

\begin{figure*}[t]
  \centering
  \includegraphics[width=\textwidth]{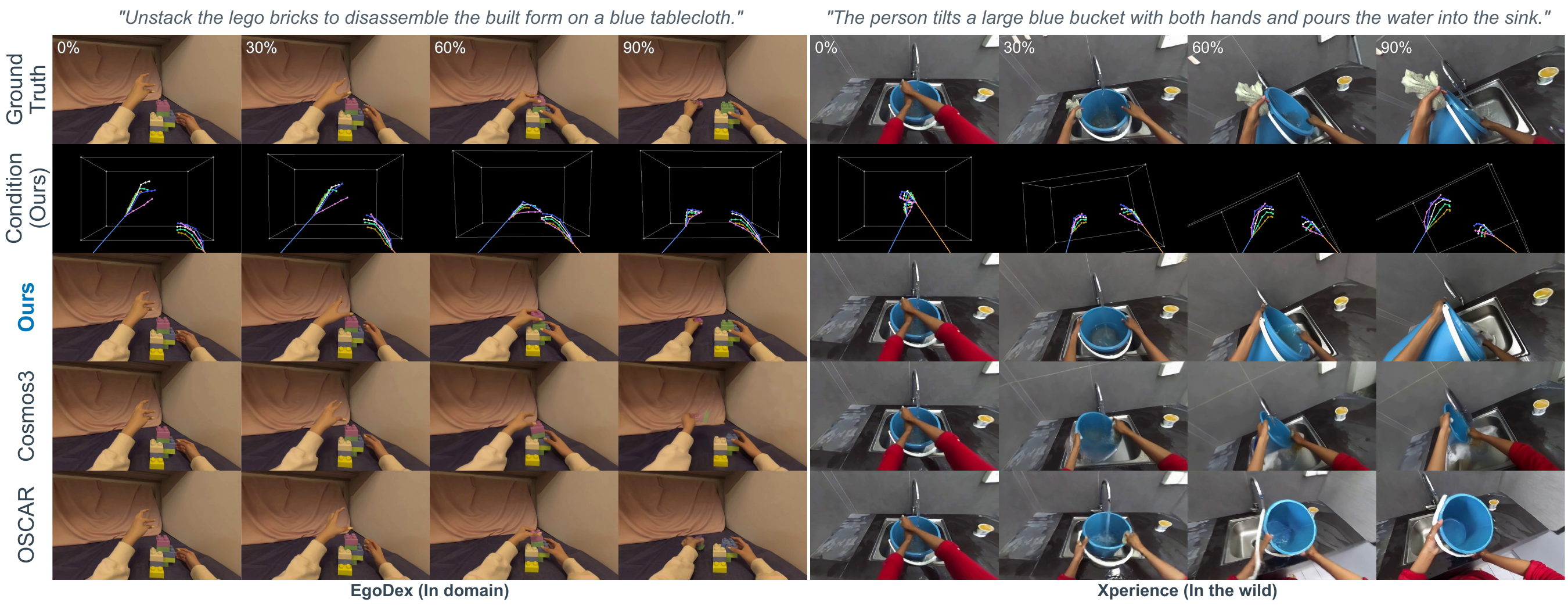}
  \caption{\textbf{Action-conditioned image-to-video generation.} Comparisons on EgoDex (left) and Xperience-10M (right, out-of-distribution).
  Adapted only on EgoDex, \paper generalizes to in-the-wild scenes.
  The generated videos follow the prescribed motion with physically coherent interactions, correctly removing the purple block (left) and tilting the bucket (right).}
  \label{fig:wm-video-comparison}
\end{figure*}

\begin{table*}[!t]
  \centering
  \fontsize{7}{8.5}\selectfont
  \setlength{\tabcolsep}{2pt}
  \caption{Action-conditioned image-to-video generation on EgoDex and Xperience-10M.
  \paper achieves strong visual fidelity and action control in both datasets.
  PCK, IoU, and Failure are in percentages; best and second-best are bold and underlined.
  EgoSim$_{\mathrm{O}}$/EgoSim$_{\mathrm{Q}}$ use dataset-provided/Qwen-generated prompts.}
  \label{tab:wm-main}
  \begin{tabular*}{\textwidth}{@{\extracolsep{\fill}}lcccccc@{\hspace{12pt}}cccccc@{}}
    \toprule
    & \multicolumn{6}{c}{EgoDex (111 clips)} & \multicolumn{6}{c}{Xperience-10M (50 clips)} \\
    \cmidrule(lr){2-7}\cmidrule(lr){8-13}
    Method & PSNR$\uparrow$ & SSIM$\uparrow$ & FVD$\downarrow$ & PCK@20$\uparrow$ & IoU$\uparrow$ & Failure$\downarrow$ & PSNR$\uparrow$ & SSIM$\uparrow$ & FVD$\downarrow$ & PCK@20$\uparrow$ & IoU$\uparrow$ & Failure$\downarrow$ \\
    \midrule
    EgoSim$_{\mathrm{O}}$~\cite{hao2026egosim} & 21.44 & 0.8572 & 333.51 & 65.11 & 63.47 & 29.26 & 16.81 & 0.6501 & 1035.55 & 47.91 & 50.96 & 56.38 \\
    LOME~\cite{gao2026lome} & 22.33 & 0.8446 & 350.61 & 60.09 & 64.87 & 37.77 & 11.99 & 0.5413 & 2114.01 & 50.85 & 57.50 & 87.24 \\
    EgoSim$_{\mathrm{Q}}$~\cite{hao2026egosim} & 22.48 & 0.8654 & 264.87 & 71.83 & 66.62 & 26.99 & 17.02 & 0.6535 & 869.67 & 51.92 & 52.41 & 55.15 \\
    Zhang et al.~\cite{zhang2026controllable} & 23.18 & 0.8429 & 253.16 & 62.78 & 64.33 & 29.14 & 14.99 & 0.5897 & 667.10 & \textbf{72.33} & \textbf{64.05} & 43.21 \\
    RynnWorld-Teleop~\cite{zhao2026rynnworldteleop} & 25.79 & 0.8796 & 258.75 & 69.28 & 71.14 & 38.43 & 16.05 & 0.6510 & 763.77 & 50.65 & 54.30 & 69.57 \\
    Cosmos3 Nano~\cite{nvidia2026cosmos3} & 26.72 & 0.8810 & 239.05 & 59.23 & 64.34 & 30.54 & \underline{18.79} & \underline{0.6932} & \underline{519.49} & 50.56 & 51.15 & 48.07 \\
    OSCAR~\cite{wu2026oscar} & 27.18 & 0.8798 & 171.85 & 85.54 & \underline{78.96} & 26.93 & 14.96 & 0.5841 & 685.74 & 61.41 & \underline{60.38} & 43.03 \\
    \midrule
    Ours w/o anchor (inference) & \underline{27.48} & \underline{0.8861} & \underline{145.69} & \textbf{86.92} & 78.57 & \underline{21.94} & 15.78 & 0.6200 & 607.23 & 56.99 & 56.26 & \underline{42.07} \\
    Ours & \textbf{30.35} & \textbf{0.9302} & \textbf{114.16} & \underline{86.52} & \textbf{79.12} & \textbf{21.69} & \textbf{19.67} & \textbf{0.7273} & \textbf{453.03} & \underline{62.24} & 57.86 & \textbf{40.66} \\
    \bottomrule
  \end{tabular*}
\end{table*}

\subsection{Implementation Details}
\label{sec:implementation-details}
We instantiate our method on the I2V and T2V variants of Wan2.2-A14B~\cite{wan2025wan} and use the EgoDex dataset~\cite{hoque2025egodex} for training.
EgoDex's controlled recordings allow us to assess generalization from visually limited training scenes to more diverse environments.
We train for one epoch on the official EgoDex training split, sampling one 41-frame clip from each episode.
Videos are processed at $832\!\times\!464$ resolution, and action renderings are generated on the fly from motion expressed in the clip's initial camera coordinate frame.
Further implementation details are provided in the supplementary material.

\subsection{Experimental Setup}
\label{sec:experimental-setup}
\paragraph{Evaluation settings.}
We evaluate the three prediction directions introduced in \cref{sec:problem-formulation} on egocentric manipulation tasks.
Quantitative experiments cover action-conditioned video generation with and without an initial image (\cref{sec:action-conditioned-video-results}), egocentric pose estimation (\cref{sec:pose-estimation-results}), and world-action policy (\cref{sec:joint-generation-results}).
Finally, we qualitatively examine text-only joint generation (\cref{sec:joint-generation-results}).
Prompts displayed in figures are abbreviated for readability.

\paragraph{Evaluation Datasets.}
The in-domain test set contains 111 clips randomly sampled from the EgoDex test split, one per activity category to ensure balanced coverage.
For out-of-distribution evaluation, we sample 50 Xperience-10M~\cite{ropedia2026xperience} clips with valid annotations, spanning diverse scenes and activities in the wild.
For quantitative world-action policy evaluation, we use HOT3D~\cite{banerjee2025hot3d}, which provides the 3D object annotations required for proximity measurement and lies outside our training domain.
We randomly sample 50 initial states with visible hands and reachable targets within 5--30\,cm, yielding 99 target-reaching tasks.
Following the camera-normalization principle of Being-H0~\cite{luo2025beingh0}, we rectify Xperience-10M and HOT3D fisheye videos to a fixed target pinhole camera and use the same rectified inputs for all methods.
Details of the benchmark construction are provided in the supplementary material.

\subsection{Action-Conditioned Ego Video Generation}
\label{sec:action-conditioned-video-results}

\paragraph{Action-conditioned image-to-video generation.}
Given an initial image, a prescribed action sequence, and a text prompt, we evaluate motion adherence and visual fidelity (\cref{eq:action-to-video-initial}).
We compare the methods in \cref{tab:wm-main} using their recommended inference settings.
Evaluation uses the first 41 frames within a common image region at $832\!\times\!480$.
We report PSNR~\cite{huynhthu2008psnr}, SSIM~\cite{wang2004ssim}, and FVD~\cite{unterthiner2018fvd} for visual quality, together with MediaPipe-based~\cite{zhang2020mediapipehands} PCK@20, hand IoU, and hand-matching failure following LOME~\cite{gao2026lome}.
Failure is the fraction of frames with detected hands in the reference video for which at least one reference hand remains unmatched in the generated video.

\Cref{tab:wm-main} shows that \paper outperforms all external baselines across the reported EgoDex metrics.
On Xperience-10M, it achieves the best PSNR, SSIM, FVD, and failure rate, demonstrating strong generalization to out-of-distribution settings.
Zhang et al.~\cite{zhang2026controllable} obtain higher PCK and hand IoU on matched hands, whereas we achieve better overall video quality with a lower hand-matching failure rate.
Removing the 3D anchor at inference lowers Xperience-10M PSNR from 19.67 to 15.78.
\Cref{fig:wm-video-comparison} provides qualitative comparisons, with additional results in supplementary \cref{fig:wm-video-comparison-supp,fig:wm-showcase-supp}.

\begin{figure}[t]
  \centering
  \includegraphics[width=\columnwidth]{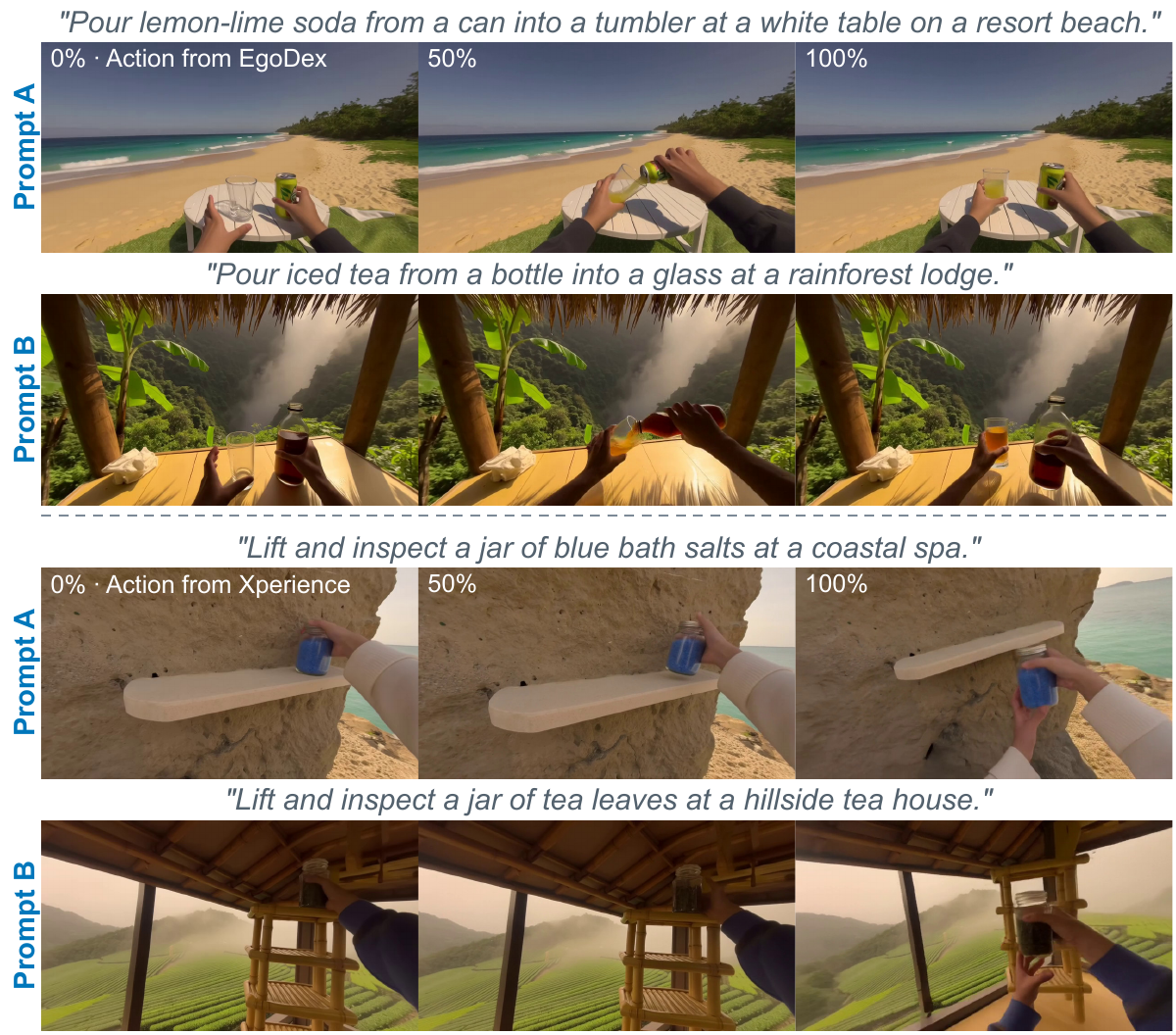}
  \caption{\textbf{Action-conditioned text-to-video generation.} \paper generates videos that share the same prescribed 3D motion while varying scenes and object appearances according to different text prompts.
  Actions are drawn from EgoDex (top) and Xperience-10M (bottom).}
  \label{fig:text-action-video}
\end{figure}

\paragraph{Action-conditioned text-to-video generation.}
Without an initial RGB image, we generate video from a prescribed action sequence and a text prompt (\cref{eq:action-to-video-full}).
For both EgoDex and Xperience-10M, we pair 20 randomly selected benchmark clips with five LLM-generated prompt variants each, yielding 100 test instances per dataset.
We compare with Wan2.2 Fun-Control and VACE-Fun~\cite{aigcapps2026videoxfun} using identical source actions and prompts.
PCK@20, hand IoU, and failure rate assess action control, while EgoVideo-S measures video--text alignment using frozen EgoVideo encoders~\cite{pei2024egovideo}, following EgoVid-5M~\cite{wang2025egovid}.
\Cref{fig:text-action-video} shows how different prompts change the visual context of a fixed action.
Additional examples and baseline comparisons are shown in supplementary \cref{fig:text-action-video-supp,fig:text-action-video-comparison-supp}.

\paper achieves the highest PCK and lowest hand-matching failure rate on both datasets (\cref{tab:text-action-video}).
Despite adaptation only on EgoDex's visually limited laboratory recordings, it remains responsive to text prompts specifying different scenes and object appearances, with comparable EgoVideo-S scores.

\begin{table}[t]
  \centering
  \fontsize{7}{8.5}\selectfont
  \setlength{\tabcolsep}{0.5pt}
  \caption{Action-conditioned text-to-video generation on 100 action--prompt instances per dataset.
  Since the varied prompts have no corresponding ground-truth videos, we evaluate motion control and video--text semantic alignment.
  Our method maintains competitive action adherence across text-specified visual contexts.}
  \label{tab:text-action-video}
  \begin{tabular*}{\columnwidth}{@{\extracolsep{\fill}}lcccc@{\hspace{4pt}}cccc@{}}
    \toprule
    & \multicolumn{4}{c}{EgoDex} & \multicolumn{4}{c}{Xperience-10M} \\
    \cmidrule(lr){2-5}\cmidrule(lr){6-9}
    Method & PCK$\uparrow$ & IoU$\uparrow$ & Fail$\downarrow$ & EgoV-S$\uparrow$ & PCK$\uparrow$ & IoU$\uparrow$ & Fail$\downarrow$ & EgoV-S$\uparrow$ \\
    \midrule
    Fun-Control & 35.41 & 48.11 & \underline{51.81} & \underline{0.3583} & 47.91 & \textbf{52.64} & \underline{63.17} & \textbf{0.4213} \\
    VACE-Fun & \underline{37.09} & \underline{48.94} & 53.74 & 0.3440 & \underline{49.04} & 49.61 & 63.33 & 0.4095 \\
    \paper & \textbf{67.91} & \textbf{66.36} & \textbf{40.15} & \textbf{0.3773} & \textbf{50.59} & \underline{50.52} & \textbf{61.99} & \underline{0.4126} \\
    \bottomrule
  \end{tabular*}
\end{table}

\subsection{Egocentric 3D Pose Estimation}
\label{sec:pose-estimation-results}
We evaluate metric hand, upper-body, and camera recovery on the same EgoDex and Xperience-10M clips used for action-conditioned video generation (\cref{eq:video-to-action}).
We compare with HaMeR~\cite{pavlakos2024hamer} and HaWoR~\cite{zhang2025hawor}, established hand reconstruction methods used for pseudo-ground-truth annotation~\cite{gavryushin2025maple,zhu2026mint}, alongside MINT~\cite{zhu2026mint}, ReViV~\cite{lyu2026reviv}, and EgoForce~\cite{millerdurai2026egoforce}.
HaMeR, HaWoR, and EgoForce receive ground-truth camera intrinsics.
HaMeR and EgoForce additionally use HaWoR's camera trajectories. 
For fair comparison, our method uses no initial action state. Results with an initial state (\cref{eq:video-to-action-initial}) appear in supplementary \cref{sec:supp-pose-initialization}.

Following SLAHMR~\cite{ye2023slahmr} and HaWoR~\cite{zhang2025hawor}, we report PA-MPJPE for local hand pose and W-/WA-MPJPE for world-space motion, as well as absolute MPJPE (Abs-MPJPE) and adjacent-frame camera errors (RPE-T and RPE-R).
Hand metrics evaluate hands appearing in the image; failure rates account for missing predictions.
Since skeletal definitions differ, body evaluation uses shared upper-limb joints.

\begin{table*}[!t]
  \centering
  \fontsize{6.7}{8.3}\selectfont
  \setlength{\tabcolsep}{0pt}
  \renewcommand{\arraystretch}{1.04}
  \caption{Hand, upper-limb, and camera reconstruction on EgoDex and Xperience-10M.
  \paper remains competitive with specialist estimators while jointly recovering body, hand, and camera motion.
  % Data indicates whether task training includes in-the-wild data.
  HaMeR $^\dagger$ and EgoForce $^\dagger$ use HaWoR's camera estimates with camera scores reported in the HaWoR row.
  ReViV keeps its native two-second input context and all methods are evaluated on 41 frames.}
  \label{tab:pose-main}
  \resizebox{\textwidth}{!}{\scriptsize Abs$\downarrow$/PA$\downarrow$/W$\downarrow$/WA$\downarrow$: absolute/Procrustes-/world-/world-aligned MPJPE (mm); RPE-T$\downarrow$ (mm), RPE-R$\downarrow$ ($^\circ$), Fail$\downarrow$ (\%): \textcolor{green!40!black}{no failure} / \textcolor{yellow!45!orange!80!black}{some fail} / \textcolor{red!75!black}{highest failure rate}.}\par
  \vspace{2pt}
  \begin{tabular*}{\textwidth}{@{}p{28pt}p{47pt}>{\centering\arraybackslash}p{29pt}>{\centering\arraybackslash}p{43pt}>{\centering\arraybackslash}p{32pt}@{\hspace{3pt}}|@{\hspace{3pt}}*{7}{>{\centering\arraybackslash}p{\dimexpr(\textwidth-195pt-\arrayrulewidth)/14\relax}}@{\hspace{10pt}}*{7}{>{\centering\arraybackslash}p{\dimexpr(\textwidth-195pt-\arrayrulewidth)/14\relax}}@{}}
    \toprule
    & & \multicolumn{3}{c@{\hspace{3pt}}|@{\hspace{3pt}}}{\textit{Method properties}} & \multicolumn{7}{c}{\textbf{EgoDex} (111 clips)} & \multicolumn{7}{c}{\textbf{Xperience-10M} (50 clips)} \\
    \cmidrule(lr){3-5}\cmidrule(lr){6-12}\cmidrule(lr){13-19}
    & Method & Camera & Pose output & Data & Abs & PA & W & WA & Fail & \mbox{RPE-T} & \mbox{RPE-R} & Abs & PA & W & WA & Fail & \mbox{RPE-T} & \mbox{RPE-R} \\
    \midrule
    \multirow{6}{*}{\shortstack[l]{\textbf{(a)}\\\textbf{Hand +}\\\textbf{camera}}} & ReViV~\cite{lyu2026reviv} & Predicted & Hand + body & Incl. wild & 334.04 & 21.21 & 126.84 & 43.96 & \textcolor{green!40!black}{0.00} & 1.08 & \underline{0.135} & 192.79 & 15.93 & 114.81 & 44.23 & \textcolor{green!40!black}{0.00} & 4.32 & 0.431 \\
     & MINT~\cite{zhu2026mint} & Predicted & Hand only & Incl. wild & 186.62 & \underline{10.50} & \underline{30.94} & \underline{18.22} & \textcolor{yellow!45!orange!80!black}{0.34} & \underline{0.74} & \textbf{0.087} & 131.97 & \underline{7.90} & \textbf{61.55} & \textbf{26.56} & \textcolor{yellow!45!orange!80!black}{0.85} & \textbf{3.09} & \underline{0.220} \\
     & HaMeR~\cite{pavlakos2024hamer}$^\dagger$ & SLAM$^\dagger$ & Hand only & Incl. wild & 119.47 & 12.12 & 77.05 & 32.31 & \textcolor{yellow!45!orange!80!black}{14.19} & --- & --- & 124.92 & \textbf{6.95} & 82.80 & 34.20 & \textcolor{yellow!45!orange!80!black}{6.79} & --- & --- \\
     & EgoForce~\cite{millerdurai2026egoforce}$^\dagger$ & SLAM$^\dagger$ & Hand + forearm & Lab only & 114.71 & 11.79 & 61.42 & 31.73 & \textcolor{red!75!black}{15.17} & --- & --- & \underline{105.50} & 8.27 & 81.13 & 39.16 & \textcolor{red!75!black}{14.24} & --- & --- \\
     & HaWoR~\cite{zhang2025hawor} & SLAM & Hand only & Lab only & \underline{111.98} & 11.75 & 46.44 & 25.09 & \textcolor{yellow!45!orange!80!black}{9.65} & 5.17 & 0.176 & 110.43 & 8.38 & 77.56 & \underline{32.71} & \textcolor{yellow!45!orange!80!black}{2.54} & 4.78 & \textbf{0.134} \\
     & \paper & Predicted & Hand + body & Lab only & \textbf{27.73} & \textbf{7.47} & \textbf{22.65} & \textbf{13.08} & \textcolor{green!40!black}{0.00} & \textbf{0.67} & 0.138 & \textbf{91.01} & 12.42 & \underline{68.23} & 33.31 & \textcolor{green!40!black}{0.00} & \underline{3.49} & 0.610 \\
    \bottomrule
  \end{tabular*}
  \par\vspace{3pt}
  \begin{tabular*}{\textwidth}{@{}p{28pt}p{47pt}>{\centering\arraybackslash}p{104pt}@{\hspace{3pt}}|@{\hspace{3pt}}*{5}{>{\centering\arraybackslash}p{\dimexpr(\textwidth-195pt-\arrayrulewidth)/10\relax}}@{\hspace{10pt}}*{5}{>{\centering\arraybackslash}p{\dimexpr(\textwidth-195pt-\arrayrulewidth)/10\relax}}@{}}
    & Method & Evaluated joints & Abs & PA & W & WA & Fail & Abs & PA & W & WA & Fail \\
    \midrule
    \multirow{5}{*}{\shortstack[l]{\textbf{(b)}\\\textbf{Upper}\\\textbf{limbs}}} & ReViV~\cite{lyu2026reviv} & \multirow{3}{*}{\shortstack{4 joints\\elbows, wrists}} & 343.90 & 95.15 & 131.08 & 106.58 & \textcolor{green!40!black}{0.00} & 382.55 & 68.94 & 143.35 & 100.07 & \textcolor{green!40!black}{0.00} \\
     & EgoForce~\cite{millerdurai2026egoforce}$^\dagger$ &  & \underline{175.79} & \underline{47.46} & \underline{113.08} & \underline{83.83} & \textcolor{red!75!black}{11.62} & \textbf{144.50} & \underline{53.57} & \underline{125.89} & \underline{89.77} & \textcolor{red!75!black}{6.93} \\
     & \paper &  & \textbf{33.25} & \textbf{15.91} & \textbf{24.28} & \textbf{19.49} & \textcolor{green!40!black}{0.00} & \underline{161.37} & \textbf{33.63} & \textbf{72.40} & \textbf{49.97} & \textcolor{green!40!black}{0.00} \\
    \cmidrule(lr){2-3}\cmidrule(lr){4-8}\cmidrule(lr){9-13}
     & ReViV~\cite{lyu2026reviv} & \multirow{2}{*}{\shortstack{6 joints\\shoulders, elbows, wrists}} & \underline{288.73} & \underline{96.05} & \underline{120.95} & \underline{107.88} & \textcolor{green!40!black}{0.00} & \underline{337.62} & \underline{84.09} & \underline{130.93} & \underline{103.55} & \textcolor{green!40!black}{0.00} \\
     & \paper &  & \textbf{32.64} & \textbf{18.44} & \textbf{24.28} & \textbf{20.98} & \textcolor{green!40!black}{0.00} & \textbf{191.19} & \textbf{55.94} & \textbf{79.84} & \textbf{65.21} & \textcolor{green!40!black}{0.00} \\
    \bottomrule
  \end{tabular*}
\end{table*}

\Cref{tab:pose-main} shows that \paper achieves the lowest hand and upper-limb reconstruction errors across all reported pose metrics on EgoDex.
Differences in skeletal definitions and body proportions create a substantial geometric domain gap between EgoDex and Xperience-10M.
Nevertheless, \paper maintains comparable performance on Xperience-10M, notably the lowest hand Abs-MPJPE, which measures absolute 3D joint error in a common first-frame camera coordinate system.
Furthermore, \paper shows the lowest aligned upper-limb errors on both datasets, underscoring its ability to jointly recover body motion and detailed hand poses.
Qualitative pose comparisons are provided in supplementary \cref{fig:pose-estimation-overlay-supp}.

\subsection{Joint Action--Video Generation}
\label{sec:joint-generation-results}

\begin{figure}[t]
  \centering
  \includegraphics[width=\columnwidth]{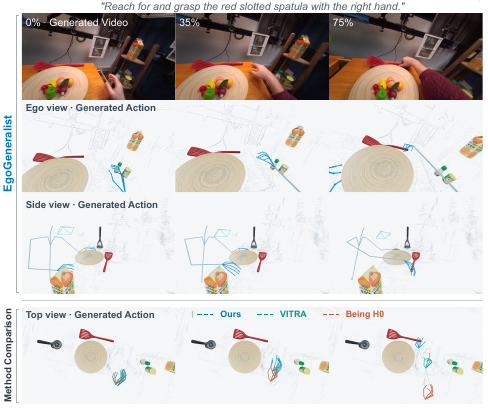}
  \caption{\textbf{World-action policy on HOT3D.} Top: \paper's generated video and 3D action in ego and side views.
  Bottom: action predictions from all three methods in a shared top view.
  GT object meshes and scene points are used only for visualization.
  Additional examples are provided in supplementary \cref{fig:wam-action-comparison-supp}.}
  \label{fig:wam-action-comparison}
\end{figure}

\begin{figure*}[t]
  \centering
  \includegraphics[width=1\textwidth]{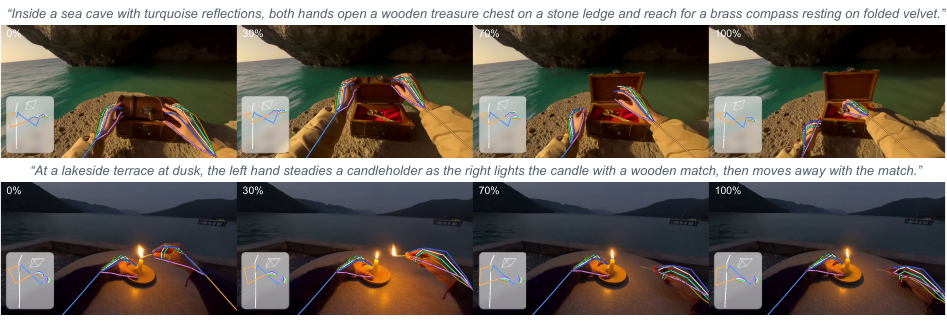}
  \caption{\textbf{Text-only joint action--video generation.} Each example shows jointly generated video and 3D action, with the action projected onto video frames.
  Insets show the generated 3D upper-body skeleton and camera pose.
  Best viewed in the supplementary video.}
  \label{fig:text-to-action-video}
\end{figure*}

\paragraph{World-action policy.}
Given an initial observation, action state, and instruction, \paper jointly predicts future 3D action and video (\cref{eq:joint-action-video-initial}).
We compare with VITRA~\cite{li2025vitra} and Being-H0~\cite{luo2025beingh0} on HOT3D~\cite{banerjee2025hot3d}, following VITRA's target-reaching criterion.
We report minimum fingertip-to-target surface distance over the predicted trajectory (MinDist), reaching rates within 2/5\,cm (Reach@2/5), and percentage distance reduction from the initial state (Progress).
\Cref{fig:wam-action-comparison} visualizes the jointly predicted video and 3D action.

\begin{table}[t]
  \centering
  \fontsize{7}{8.5}\selectfont
  \setlength{\tabcolsep}{0.5pt}
  \caption{World-action policy evaluation on HOT3D.
  Our method outperforms specialized policy baselines in target reaching on unseen HOT3D scenes while jointly generating video.
  All methods receive an initial observation and instruction.
  VITRA and \paper also receive initial action states. $^\dagger$Being-H0 uses a single wrist-anchored rigid alignment to the GT initial hand.}
  \label{tab:wam-action}
  \begin{tabular*}{\columnwidth}{@{\extracolsep{\fill}}lccccc@{}}
    \toprule
    & \multicolumn{2}{c}{MinDist (cm)$\downarrow$} & \multicolumn{2}{c}{Reach (\%)$\uparrow$} & Progress$\uparrow$ \\
    \cmidrule(lr){2-3}\cmidrule(lr){4-5}
    Method & Mean & Med. & @2cm & @5cm & (\%) \\
    \midrule
    Initial state & 15.98 & 16.20 & 0.00 & 0.00 & 0.00 \\
    \midrule
    Being-H0$^\dagger$ & 11.64 & 11.90 & 16.16 & 25.25 & 32.64 \\
    VITRA & 6.32 & 4.59 & 31.31 & 52.53 & 61.96 \\
    \paper & \textbf{3.81} & \textbf{1.07} & \textbf{58.59} & \textbf{77.78} & \textbf{77.42} \\
    \bottomrule
  \end{tabular*}
\end{table}

As shown in \cref{tab:wam-action}, our method achieves lower mean and median minimum target distances and higher reaching rates.
Despite training only on EgoDex, the generated actions follow target-specific instructions for unseen objects and environments.

\paragraph{Text-only joint action--video generation.}
Given only a text prompt, \paper generates synchronized video and metric 3D motion (\cref{eq:joint-action-video-full}).
\Cref{fig:text-to-action-video} illustrates scenes visually distinct from the training domain.
The generated actions remain closely aligned with the motion depicted in the videos.
These qualitative results illustrate the potential for paired action--video data generation beyond the training settings.
More examples are in the supplementary material.

% !TeX root = ../main.tex
\subsection{Ablation Studies}
\label{sec:rendering-ablation}
To evaluate the effectiveness of role-adaptive rendering, we compare ego-view rendering and multiview \emph{Posegrid} for both video control and metric action readout.
We further assess cross-stream action--video coupling against alternative conditioning strategies.
These ablations validate our designs.
Detailed ablations and results are presented in \cref{sec:supp-ablation} of the supplementary material.

% !TeX root = ../main.tex

\section{Conclusion}
\label{sec:conclusion}

We presented \paper, a unified framework for action-conditioned ego video generation, metric pose estimation, and joint action--video generation.
Role-adaptive rendering and cross-stream coupling connect metric motion with the video generator's native representation through lightweight adaptation.
Experiments demonstrate effective action control and generalization beyond laboratory training scenes, alongside competitive motion recovery and target-reaching performance.
Text-only joint generation further enables paired synthesis of action and video.
Together, these results establish pretrained video models as a practical foundation for egocentric action--video learning.

\paragraph{Limitations.}
EgoDex's multi-camera fusion introduces projection mismatches between 3D annotations and images.
These discrepancies can carry over into generated sequences, occasionally affecting action--video alignment.
More accurately aligned annotations could reduce these artifacts.
Furthermore, although our method generalizes well, its training data primarily cover tabletop manipulation.
Including more varied body postures and interaction settings could broaden our framework's capabilities.

{
    \small
    \bibliographystyle{ieeenat_fullname}
    \bibliography{main}
}

% WARNING: do not forget to delete the supplementary pages from your submission 
% !TeX root = ../main.tex
\clearpage
\hypersetup{hypertexnames=false}
\makeatletter
\global\Hy@pagecounter=\numexpr\value{page}-1\relax
\makeatother
\setcounter{page}{1}
\appendix
\maketitlesupplementary
\setcounter{figure}{0}
\renewcommand{\thefigure}{S\arabic{figure}}
\renewcommand{\theHfigure}{supp.\arabic{figure}}

\section{Implementation Details}
\label{sec:supp-implementation}

\paragraph{Initial-state conditioning.}
Our framework allows all tasks to share a single backbone.
In our implementation, however, we use separate Wan2.2 I2V and T2V checkpoints to match their native conditioning settings and better leverage their pretrained generative priors.
The I2V variant retains the backbone's native first-frame conditioning for the target streams, following the task definitions in \cref{sec:problem-formulation}.
The T2V variant generates the complete target sequence without an initial target state.

\paragraph{Optimization.}
We train the two variants independently, sampling the three prediction directions uniformly (1:1:1).
The pretrained backbone weights and video VAE remain frozen, and all LoRA adapters use rank 32.
Attention LoRAs adapt the query, key, value, and output projections of self-attention.
We use AdamW~\cite{loshchilov2019adamw} with an initial learning rate of $1\!\times\!10^{-4}$, cosine learning-rate decay, weight decay of $0.01$, and gradient clipping at $1.0$.
Training runs on NVIDIA A800 GPUs with BF16 mixed precision and a global batch size of 16.

\paragraph{Inference.}
We use 40 denoising steps.
For text-conditioned generation, we follow the default classifier-free guidance settings of the corresponding Wan2.2 backbone.
Pose estimation does not use text conditioning.

\subsection{Posegrid Decoder}
\label{sec:supp-posegrid-decoder}

\paragraph{Architecture.}
The \emph{Posegrid} decoder converts latents from the frozen video VAE into metric 3D motion through a shared spatiotemporal encoder and two output branches.
The encoder projects the 16-channel latents to 384 channels and processes them with three residual 3D convolutional blocks.
Features are spatially pooled to a $12\times16$ grid, projected to a temporal sequence, and refined by four residual temporal blocks.
Linear interpolation restores the original frame count.
The two output branches recover positions and rotations for 68 skeletal joints and the camera, all expressed in the initial camera coordinate frame.
We use replicate padding in the convolutional blocks to avoid introducing zero-valued features at sequence boundaries.

\paragraph{Training.}
We train the decoder independently on the official EgoDex training split to reconstruct metric motion from ground-truth \emph{Posegrid} latents produced by the frozen video VAE.
The reconstruction objective combines joint and camera supervision:
\begin{equation}
\label{eq:supp-decoder-loss}
\mathcal{L}_{\mathrm{dec}}
=
\sum_{s\in\{\mathrm{joint},\,\mathrm{camera}\}}
\left(
\mathcal{L}_{\mathrm{pos}}^{s}
+0.3\,\mathcal{L}_{\mathrm{rot}}^{s}
+0.5\,\mathcal{L}_{\mathrm{disp}}^{s}
\right).
\end{equation}
Here, $s$ indexes joint or camera predictions.
$\mathcal{L}_{\mathrm{pos}}^{s}$ and $\mathcal{L}_{\mathrm{rot}}^{s}$ are mean squared errors in 3D positions and 6D rotation representations, respectively.
$\mathcal{L}_{\mathrm{disp}}^{s}$ supervises adjacent-frame position differences to encourage temporal consistency.
Joint losses assign weights of 1, 5, and 10 to body joints, hand joints, and fingertips, respectively, with invalid annotations masked.
We optimize only decoder parameters using AdamW with an initial learning rate of $10^{-4}$, cosine decay toward $10^{-6}$, weight decay of $0.01$, a global batch size of 32, and gradient clipping at $1.0$.
After training, the decoder is frozen and maps generated \emph{Posegrid} latents directly to metric motion.

\section{Benchmark Construction and Evaluation Protocols}
\label{sec:supp-benchmark-protocols}

\subsection{Dataset Construction}

\paragraph{EgoDex.}
We construct the in-domain benchmark from the official EgoDex test split, randomly selecting one episode from each of its 111 activity categories.
Evaluation uses the first 41 frames of each selected episode, with identical frame indices for all methods.
Action-conditioned video generation and pose estimation share these 111 clips.

\paragraph{Xperience-10M.}
The out-of-distribution benchmark contains 50 clips spanning diverse scenes and activities, such as plant watering, cooking, household chores, and shopping.
The released sample data contain many low-quality segments and an overrepresentation of repetitive craft activities.
We therefore automatically exclude segments with incomplete annotations and filter tasks by their titles, excluding assembly- and bead-related tasks, along with repetitive paper crafts and cardboard-cutting tasks.
We then screen out locomotion-dominated segments to focus evaluation on manipulation.
Finally, we randomly sample 50 clips from the remaining candidates to cover diverse scene types.
All methods are evaluated on the same selected clips and identical 41-frame windows.

\paragraph{HOT3D.}
We construct the world-action policy benchmark from 136 HOT3D Aria recordings with ground-truth annotations.
Candidate initial states are screened at two-second intervals for annotation validity, hand visibility, non-contact initialization, and target reachability, with redundant nearby states excluded.
Eligible targets have reliable geometry and sufficient visibility, lie within 60 cm of the camera, and are 5--30 cm from an operating fingertip.
We randomly sample from the eligible candidates and the final benchmark contains 50 initial states and 99 tasks, each specifying an initial state, target object, and operating hand.
We complete the missing HOT3D body state using EgoDex pose and bone-length statistics, providing an approximate initialization that need not be accurate.

\paragraph{Action-conditioned text-to-video generation.}
We randomly select 20 activity categories from the fixed EgoDex benchmark and 20 clips from the fixed Xperience-10M benchmark.
An LLM generates five prompt variants per source caption using a fixed instruction that preserves the action, its order, and explicitly specified hand roles while varying compatible environments and object appearance.
This produces 100 test instances per dataset.
All methods receive the same text and source motion, converted into their supported conditioning representations.

We use the following prompt-generation instruction, replacing ``EgoDex'' with ``Xperience'' for the corresponding dataset:
\begin{quote}
\textit{``For each provided EgoDex test prompt, generate 5 variations.
Structure each prompt as an action description specifying the objects and hand roles, followed by a concise description of the setting and lighting.
Preserve the original action and action order.
Vary the environment, lighting, and object appearance while keeping them compatible with the original interaction.
Use diverse, plausible indoor and outdoor settings with clear visual details.''}
\end{quote}

\subsection{Metric Computation}

\paragraph{Video generation.}
We compute PSNR and SSIM per frame and average the scores first within clips and then across clips.
For hand alignment, MediaPipe detects joints in both reference and generated videos.
PCK@20 measures the fraction of matched joints within 20 pixels of their corresponding reference detections, while hand IoU measures the intersection over union of matched hand bounding boxes.
Both metrics are computed over matched hands.
Failure is the fraction of frames with at least one unmatched reference hand among frames containing reference-hand detections.
EgoVideo-S measures cosine similarity between frozen text and video embeddings, using frames 5, 15, 25, and 35.

\paragraph{Pose reconstruction.}
Abs-MPJPE measures mean Euclidean joint error in a common first-frame camera coordinate system, preserving the predicted scale.
PA-MPJPE independently aligns each hand to its reference through a per-frame similarity transform.
Following HaWoR~\cite{zhang2025hawor} and SLAHMR~\cite{ye2023slahmr}, W-MPJPE fits one similarity transform per hand using frames 0--1 and applies it throughout the clip.
WA-MPJPE instead fits the transform using the complete clip.
For upper-limb evaluation, both arms form a single alignment group.
Reconstruction errors are computed over valid prediction--reference pairs, while failure statistics account for missing pose predictions and unavailable camera estimates required for world-space reconstruction.
Camera RPE-T and RPE-R measure translation and rotation errors in adjacent-frame relative motion.

\paragraph{Target reaching.}
For each task, let $d_0$ denote the minimum initial fingertip-to-target distance and $d_{\min}$ the minimum over generated states.
Distances are computed to mesh triangles with the target fixed at its initial pose.
Reach@2/5 reports the percentage of tasks with $d_{\min}$ within 2/5 cm, and Progress is $100\max(0,1-d_{\min}/d_0)$.
Results are aggregated over individual tasks.

% !TeX root = ../main.tex
\section{Ablation Experiments}
\label{sec:supp-ablation}

To limit computational cost, we train on smaller EgoDex subsets for fewer updates.
Within each ablation group, variants use the same training data and number of updates and are evaluated at the end of training.
We use the EgoDex and Xperience-10M evaluation sets from the main experiments.

\begin{table}[t]
  \centering
  \fontsize{7}{8.5}\selectfont
  \setlength{\tabcolsep}{1pt}
  \caption{\textbf{Hand, upper-limb, and camera readout on EgoDex.} Multiview Posegrid gives lower reconstruction errors, indicating that its VAE latents better preserve metric action information for readout.}
  \label{tab:rendering-ablation}
  \begin{tabular*}{\columnwidth}{@{\extracolsep{\fill}}lcccccc@{}}
    \toprule
    \multicolumn{7}{l}{(a) Hands and camera} \\
    Rendering & Abs$\downarrow$ & PA$\downarrow$ & W$\downarrow$ & WA$\downarrow$ & RPE-T$\downarrow$ & RPE-R$\downarrow$ \\
    \midrule
    Ego-view & 38.70 & 12.61 & 36.28 & 22.07 & 2.13 & 0.219 \\
    Posegrid & \textbf{21.67} & \textbf{12.01} & \textbf{22.91} & \textbf{15.74} & \textbf{1.59} & \textbf{0.186} \\
    \bottomrule
  \end{tabular*}
  \par\smallskip
  \begin{tabular*}{\columnwidth}{@{\extracolsep{\fill}}lcccc@{}}
    \toprule
    \multicolumn{5}{l}{(b) Upper limbs (six joints)} \\
    Rendering & Abs$\downarrow$ & PA$\downarrow$ & W$\downarrow$ & WA$\downarrow$ \\
    \midrule
    Ego-view & 53.68 & 32.49 & 44.86 & 38.38 \\
    Posegrid & \textbf{38.66} & \textbf{26.04} & \textbf{32.34} & \textbf{28.95} \\
    \bottomrule
  \end{tabular*}
\end{table}

\begin{table*}[t]
  \centering
  \fontsize{7}{8.5}\selectfont
  \setlength{\tabcolsep}{1.5pt}
  \caption{\textbf{Ablations of two design components for action-conditioned video generation.} We evaluate how different action--video coupling strategies and action renderings affect video controllability under matched data and training budgets. The full configuration performs best on most metrics, supporting our cross-stream coupling and role-adaptive rendering design. Best and second-best scores are bold and underlined.}
  \label{tab:coupling-ablation}
  \begin{tabular*}{\textwidth}{@{\extracolsep{\fill}}llcccccc@{\hspace{8pt}}cccccc@{}}
    \toprule
    & & \multicolumn{6}{c}{EgoDex (111 clips)} & \multicolumn{6}{c}{Xperience-10M (50 clips)} \\
    \cmidrule(lr){3-8}\cmidrule(lr){9-14}
    Ablation component & Configuration & PSNR$\uparrow$ & SSIM$\uparrow$ & FVD$\downarrow$ & PCK@20$\uparrow$ & IoU$\uparrow$ & Failure$\downarrow$ & PSNR$\uparrow$ & SSIM$\uparrow$ & FVD$\downarrow$ & PCK@20$\uparrow$ & IoU$\uparrow$ & Failure$\downarrow$ \\
    \midrule
    \multirow{2}{*}{Action--video coupling} & Action embedding & 22.14 & 0.834 & 369.06 & 31.94 & 43.16 & 48.91 & 14.72 & 0.589 & 790.48 & 25.65 & 34.19 & 64.80 \\
    & Latent channel concatenation & 24.02 & 0.854 & 270.98 & 43.23 & 52.87 & 34.57 & 15.06 & 0.605 & 823.82 & 27.94 & 37.76 & 57.29 \\
    \midrule
    \multirow{2}{*}{Action rendering} & Multiview Posegrid & 24.95 & 0.865 & 254.77 & 50.53 & 58.21 & 33.56 & 15.01 & 0.604 & 749.33 & 36.73 & 43.36 & 58.19 \\
    & Ego-view w/o 3D anchor & \underline{26.07} & \underline{0.877} & \underline{205.15} & \underline{67.09} & \underline{69.37} & \underline{24.71} & \underline{15.65} & \underline{0.626} & \underline{667.88} & \underline{58.03} & \textbf{57.67} & \textbf{38.45} \\
    \midrule
    Full model & Ego-view + 3D anchor (ours) & \textbf{28.42} & \textbf{0.916} & \textbf{170.96} & \textbf{68.43} & \textbf{71.46} & \textbf{24.57} & \textbf{19.06} & \textbf{0.707} & \textbf{506.83} & \textbf{58.77} & \underline{56.79} & \underline{41.18} \\
    \bottomrule
  \end{tabular*}
\end{table*}

\subsection{Role-Adaptive Rendering}
\label{sec:supp-rendering-ablation}
\begin{figure}[t]
  \centering
  \includegraphics[width=\columnwidth]{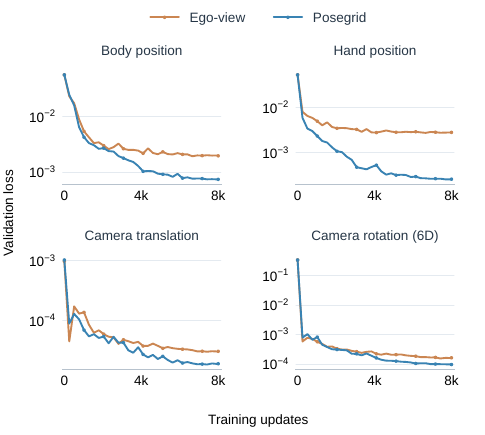}
  \caption{\textbf{Decoder learning curves.} Multiview Posegrid enables faster loss reduction, particularly for body and hand positions, and reaches lower final validation losses across the four reported components, making it better suited for metric action readout.}
  \label{fig:decoder-readout-loss}
\end{figure}

We test whether multiview \emph{Posegrid} is better suited to metric readout and ego-view rendering to video control.
For readout, we train identical decoders on GT ego-view renderings and multiview \emph{Posegrid}, encoded by the same frozen VAE.
Both receive body, hand, and camera supervision and use the same 8,192 training episodes for 8,000 updates at global batch size 32.
\Cref{tab:rendering-ablation} reports hand, upper-limb, and camera reconstruction errors on EgoDex.
Multiview \emph{Posegrid} produces lower hand, upper-limb, and camera reconstruction errors, supporting its use for metric action readout.
\Cref{fig:decoder-readout-loss} presents the decoder validation curves.

For control, we compare three action renderings: multiview \emph{Posegrid}, ego-view rendering without the 3D anchor cage, and our full ego-view rendering with the cage (\cref{tab:coupling-ablation}, ``Action rendering'' and ``Full model'').
Unlike the inference-time removal of the 3D anchor cage in \cref{tab:wm-main}, the no-anchor variant here is trained and evaluated without the 3D anchor cage.
All variants in \cref{tab:coupling-ablation} use 2,048 training episodes and 1,200 updates at a global batch size of 8, and are evaluated using the same video-generation metrics.
Ego-view rendering with the 3D anchor cage performs best on most metrics, followed by its no-anchor variant, whereas multiview \emph{Posegrid} is less effective for video control.
The no-anchor variant obtains higher hand IoU and a lower failure rate on Xperience-10M, but the lack of explicit camera control leads to worse visual metrics than the full method.

\subsection{Cross-Stream Action--Video Coupling}
\label{sec:supp-coupling-ablation}
Using the video-generation training and evaluation settings in \cref{sec:supp-rendering-ablation}, we further evaluate the advantages of cross-stream action--video coupling design.
We compare cross-stream coupling with two alternative conditioning strategies: \emph{latent channel concatenation}, which concatenates action and video latents along the channel dimension, and \emph{action embedding}, which injects encoded numerical motion through separate cross-attention.
All variants train the active LoRA adapters, together with any added conditioning modules.
Our full configuration outperforms both alternative coupling strategies across all reported metrics on both datasets (\cref{tab:coupling-ablation}), learning more effective action control and video generation within the same short training budget.
Despite using only 0.65\% of the full training episodes and limited updates, our method approaches the full-training results on several metrics (\cref{tab:wm-main}), suggesting the potential for data-efficient adaptation through effective reuse of the pretrained video prior.

\section{Additional Results}
\label{sec:supp-qualitative}

\begin{table*}[t]
  \centering
  \fontsize{7}{8.5}\selectfont
  \setlength{\tabcolsep}{1pt}
  \caption{\textbf{Pose estimation with and without an initial action state.}
  GT first-frame conditioning lowers most reconstruction errors on both datasets.
  Metrics and units follow \cref{tab:pose-main}.
  All 41 frames are evaluated. Lower errors are bold.}
  \label{tab:pose-initialization}
  \begin{tabular*}{\textwidth}{@{\extracolsep{\fill}}llcccc@{\hspace{8pt}}cccc@{\hspace{8pt}}cc@{}}
    \toprule
    & & \multicolumn{4}{c}{Hands} & \multicolumn{4}{c}{Upper limbs (six joints)} & \multicolumn{2}{c}{Camera} \\
    \cmidrule(lr){3-6}\cmidrule(lr){7-10}\cmidrule(lr){11-12}
    Dataset & Initial action state & Abs$\downarrow$ & PA$\downarrow$ & W$\downarrow$ & WA$\downarrow$ & Abs$\downarrow$ & PA$\downarrow$ & W$\downarrow$ & WA$\downarrow$ & RPE-T$\downarrow$ & RPE-R$\downarrow$ \\
    \midrule
    \multirow{2}{*}{EgoDex} & None & 27.73 & 7.47 & 22.65 & 13.08 & 32.64 & 18.44 & 24.28 & 20.98 & 0.67 & 0.138 \\
    & GT first frame & \textbf{15.22} & \textbf{6.10} & \textbf{14.12} & \textbf{10.53} & \textbf{17.90} & \textbf{11.48} & \textbf{15.53} & \textbf{13.58} & \textbf{0.63} & \textbf{0.128} \\
    \midrule
    \multirow{2}{*}{Xperience-10M} & None & 91.01 & 12.42 & 68.23 & \textbf{33.31} & 191.19 & 55.94 & 79.84 & 65.21 & 3.49 & 0.610 \\
    & GT first frame & \textbf{83.62} & \textbf{12.08} & \textbf{56.67} & 33.53 & \textbf{158.32} & \textbf{54.51} & \textbf{76.42} & \textbf{63.46} & \textbf{3.46} & \textbf{0.600} \\
    \bottomrule
  \end{tabular*}
\end{table*}

We provide a quick preview video highlighting \paper's capabilities and a supplementary video presenting detailed experimental results.

\subsection{Action-Conditioned Ego Video Generation}

\begin{figure*}[t]
  \centering
  \includegraphics[width=\textwidth]{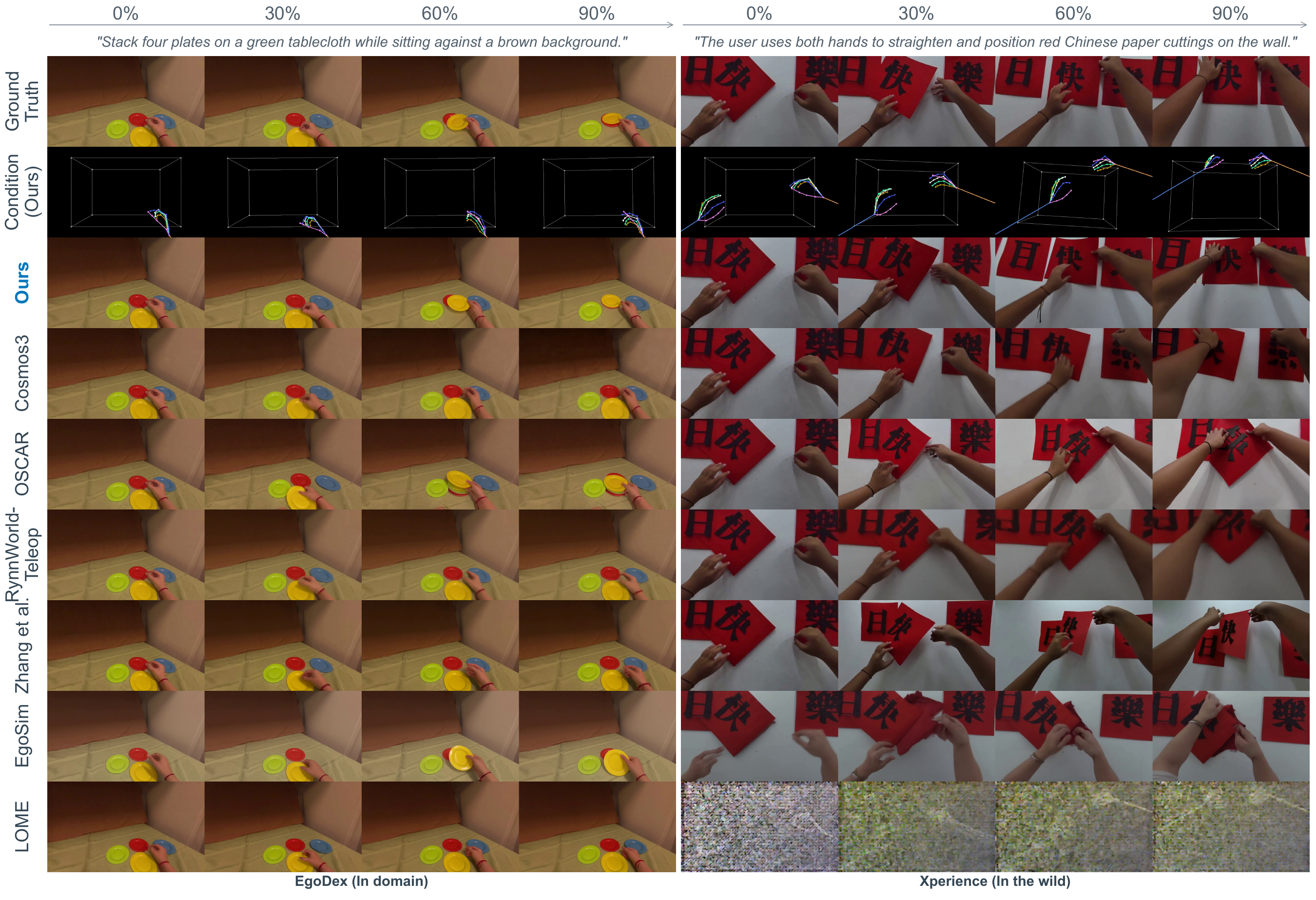}
  \caption{Additional action-conditioned image-to-video generation comparisons on EgoDex and Xperience-10M.
  \paper maintains effective action control and coherent object interactions.
  Columns show 0\%, 30\%, 60\%, and 90\% progress through each 41-frame clip.}
  \label{fig:wm-video-comparison-supp}
\end{figure*}

\begin{figure*}[t]
  \centering
  \includegraphics[width=\textwidth,height=0.92\textheight,keepaspectratio]{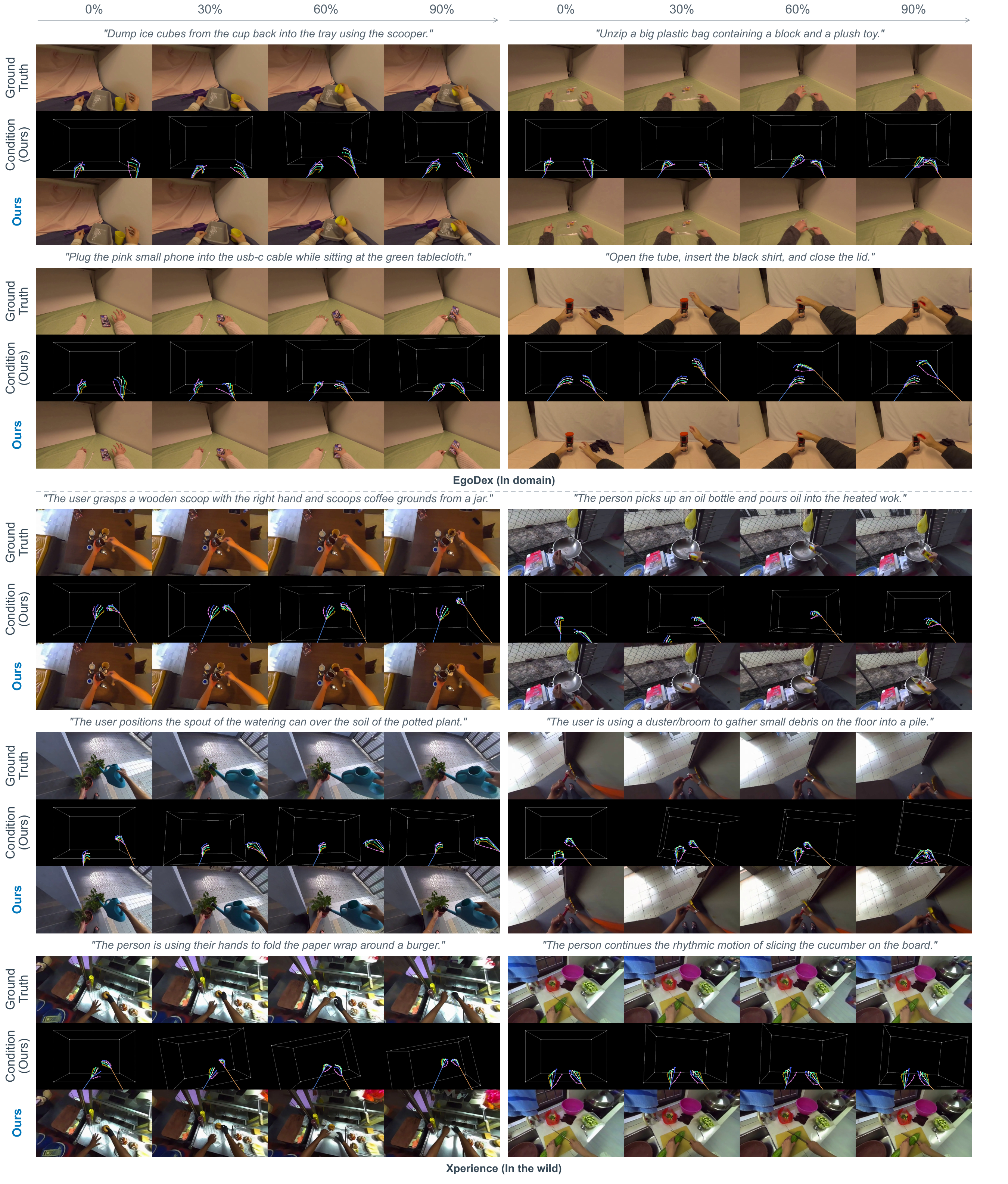}
  \caption{Action-conditioned image-to-video generation across EgoDex and Xperience-10M.
  Adapted only on EgoDex, \paper generates motion-consistent interactions across laboratory scenes and diverse in-the-wild environments.
  Each example shows reference frames, the ego-view action condition, and our generated video.
  Columns show 0\%, 30\%, 60\%, and 90\% progress through each 41-frame clip.}
  \label{fig:wm-showcase-supp}
\end{figure*}

\paragraph{Action-conditioned image-to-video generation.}
\Cref{fig:wm-video-comparison-supp,fig:wm-showcase-supp} extend the comparisons and examples on EgoDex and Xperience-10M, showing generated videos alongside ground truth and ego-view action conditions.

\begin{figure*}[t]
  \centering
  \includegraphics[width=\textwidth,height=0.90\textheight,keepaspectratio]{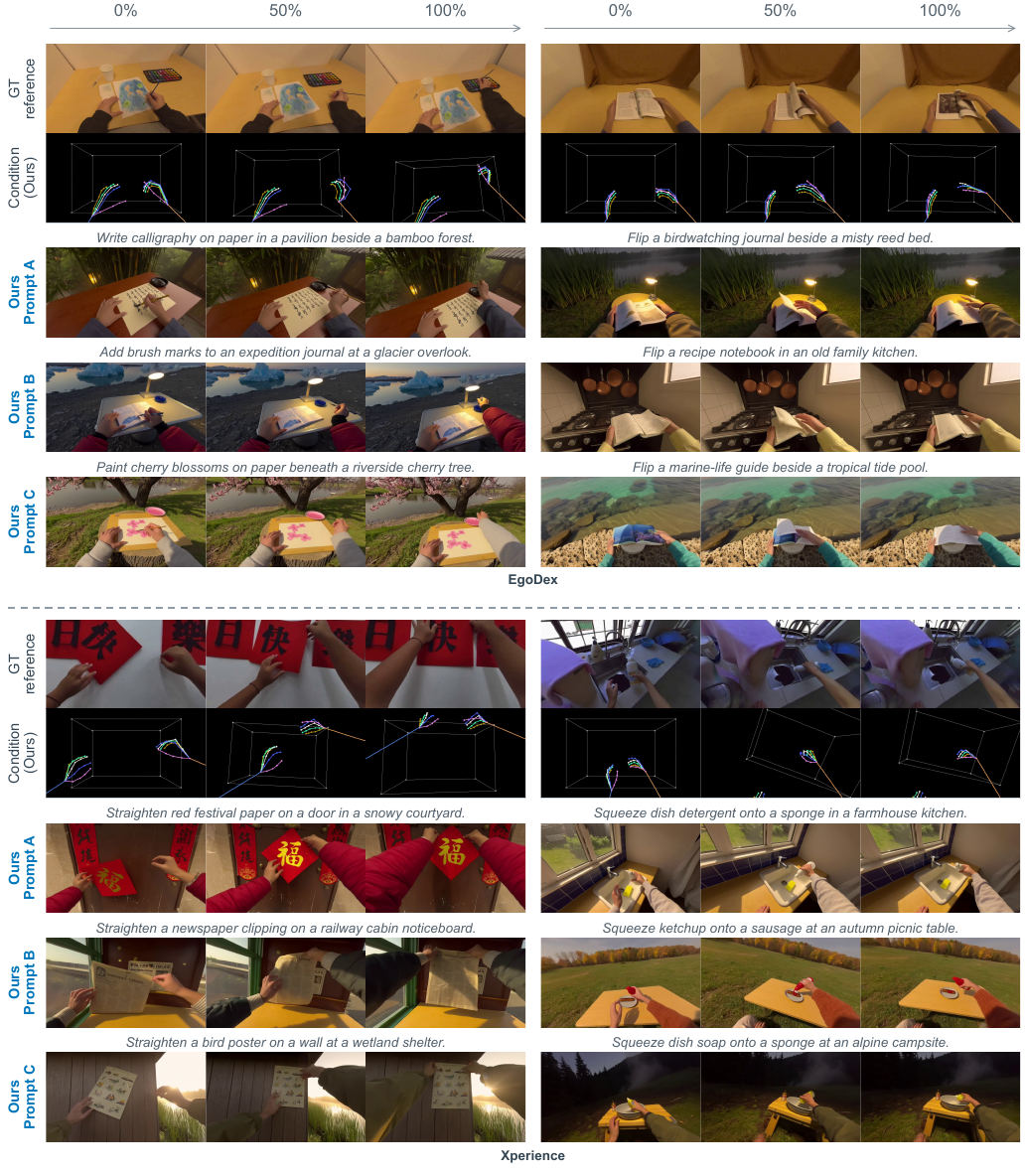}
  \caption{Additional action-conditioned text-to-video generation.
  Action sequences from EgoDex (top) and Xperience-10M (bottom) are each paired with three text prompts.
  \paper varies scenes and objects according to text while maintaining the prescribed motion.
  GT video is reference only.}
  \label{fig:text-action-video-supp}
\end{figure*}

\begin{figure*}[t]
  \centering
  \includegraphics[width=\textwidth]{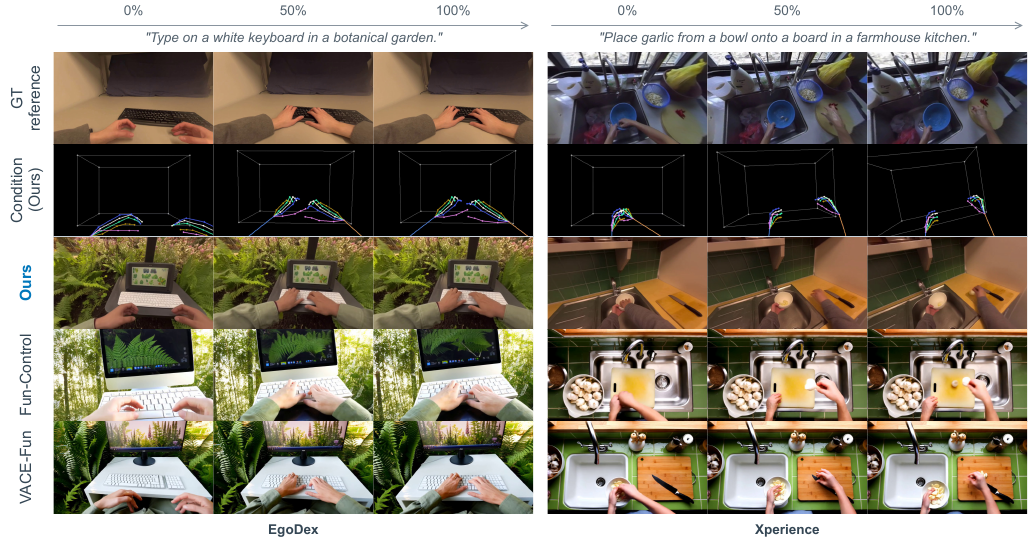}
  \caption{Action-conditioned text-to-video generation on EgoDex (left) and Xperience-10M (right), compared with Wan2.2 Fun-Control and VACE-Fun.
  Our 3D anchor enables explicit camera-motion control, allowing \paper to reproduce prescribed viewpoint changes more faithfully than the baselines.
  Methods share source actions and prompts but use their native condition representations.
  The displayed ego-view condition is ours.
  Columns show frames 0, 20, and 40 of the shared 41-frame evaluation window.}
  \label{fig:text-action-video-comparison-supp}
\end{figure*}

\paragraph{Action-conditioned text-to-video generation.}
\Cref{fig:text-action-video-supp} illustrates how different prompts vary the visual context of a prescribed action without an initial RGB image.
\Cref{fig:text-action-video-comparison-supp} compares our method with Wan2.2 Fun-Control and VACE-Fun using the same source actions and prompts.

\subsection{Egocentric 3D Pose Estimation}
\label{sec:supp-pose-initialization}

\paragraph{Initial-state conditioning.}
\Cref{tab:pose-initialization} extends the evaluation in \cref{tab:pose-main} to include pose estimation with GT initial-state conditioning (\cref{eq:video-to-action-initial}).
Initial-state conditioning lowers overall errors on both datasets.
The improvement is smaller on Xperience-10M, where differences in skeletal definitions already introduce a geometric shift in the initial target \emph{Posegrid}.

\begin{figure*}[t]
  \centering
  \includegraphics[width=\textwidth,height=0.91\textheight,keepaspectratio]{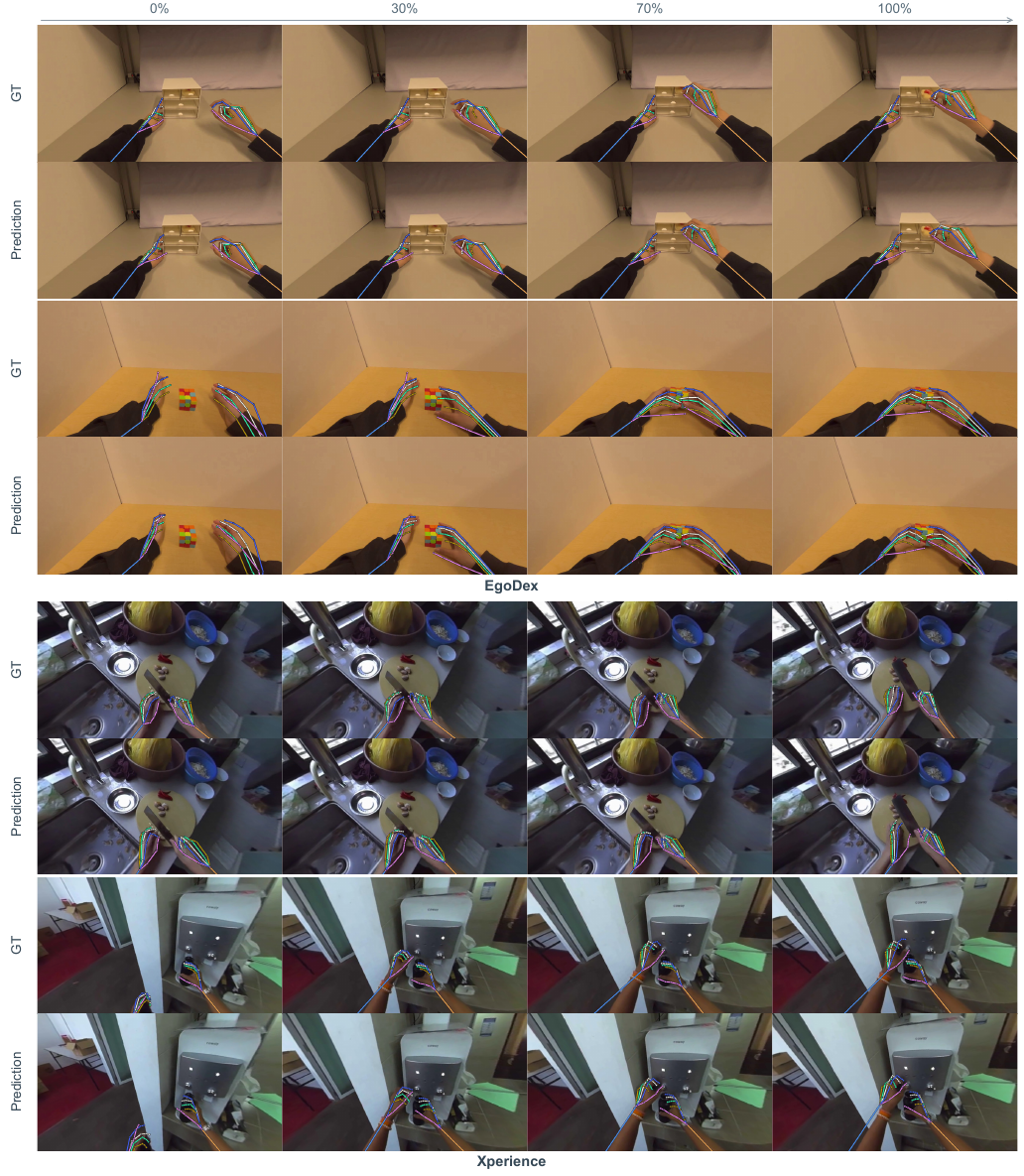}
  \caption{\textbf{Egocentric pose estimation.} Ground-truth (GT) and predicted hand and arm poses are projected onto the same input frames on EgoDex (top two examples) and Xperience-10M (bottom two examples).
  \paper receives no initial action state.
  \paper's projected hand and arm predictions align closely with the visible hands and ground-truth overlays.
  Columns indicate relative progress through each displayed window.}
  \label{fig:pose-estimation-overlay-supp}
\end{figure*}

\paragraph{Qualitative comparisons.}
\Cref{fig:pose-estimation-overlay-supp} compares ground-truth and predicted hand and upper-limb poses projected onto the input videos from EgoDex and Xperience-10M. The displayed results use no initial action state.

\subsection{Joint Action--Video Generation}

\begin{figure*}[t]
  \centering
  \includegraphics[width=\textwidth,height=0.93\textheight,keepaspectratio]{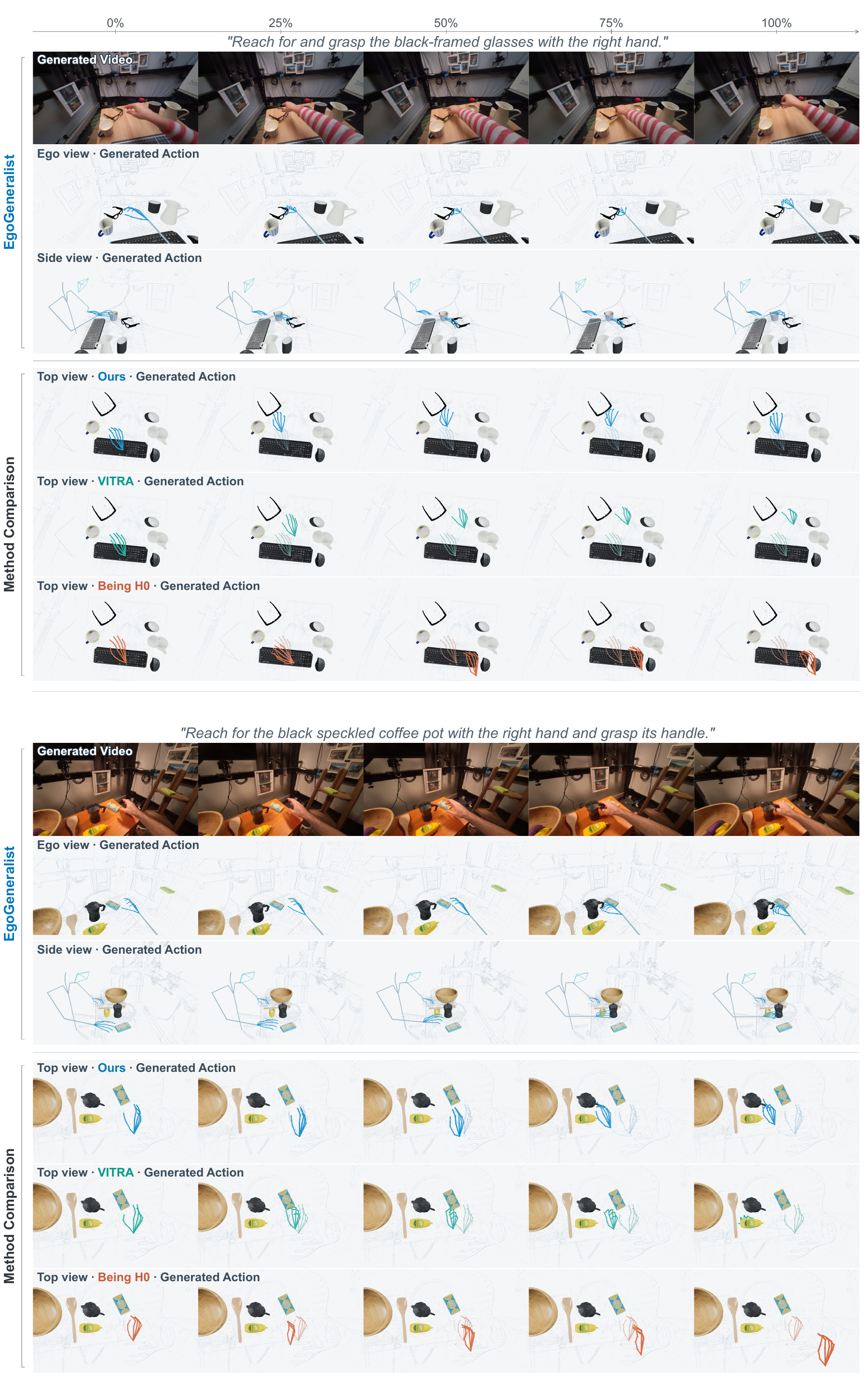}
  \caption{Additional HOT3D examples of joint action--video generation and hand-action comparisons.
  \paper generates smooth hand trajectories that accurately approach the specified objects, whereas the baseline trajectories miss the targets in these examples.
  Columns show normalized prediction progress, with visualization conventions following \cref{fig:wam-action-comparison}.}
  \label{fig:wam-action-comparison-supp}
\end{figure*}

\paragraph{World-action policy.}
\Cref{fig:wam-action-comparison-supp} presents additional HOT3D examples, showing jointly generated video and action together with hand-trajectory comparisons against the policy baselines.

\begin{figure*}[t]
  \centering
  \includegraphics[width=\textwidth,height=0.93\textheight,keepaspectratio]{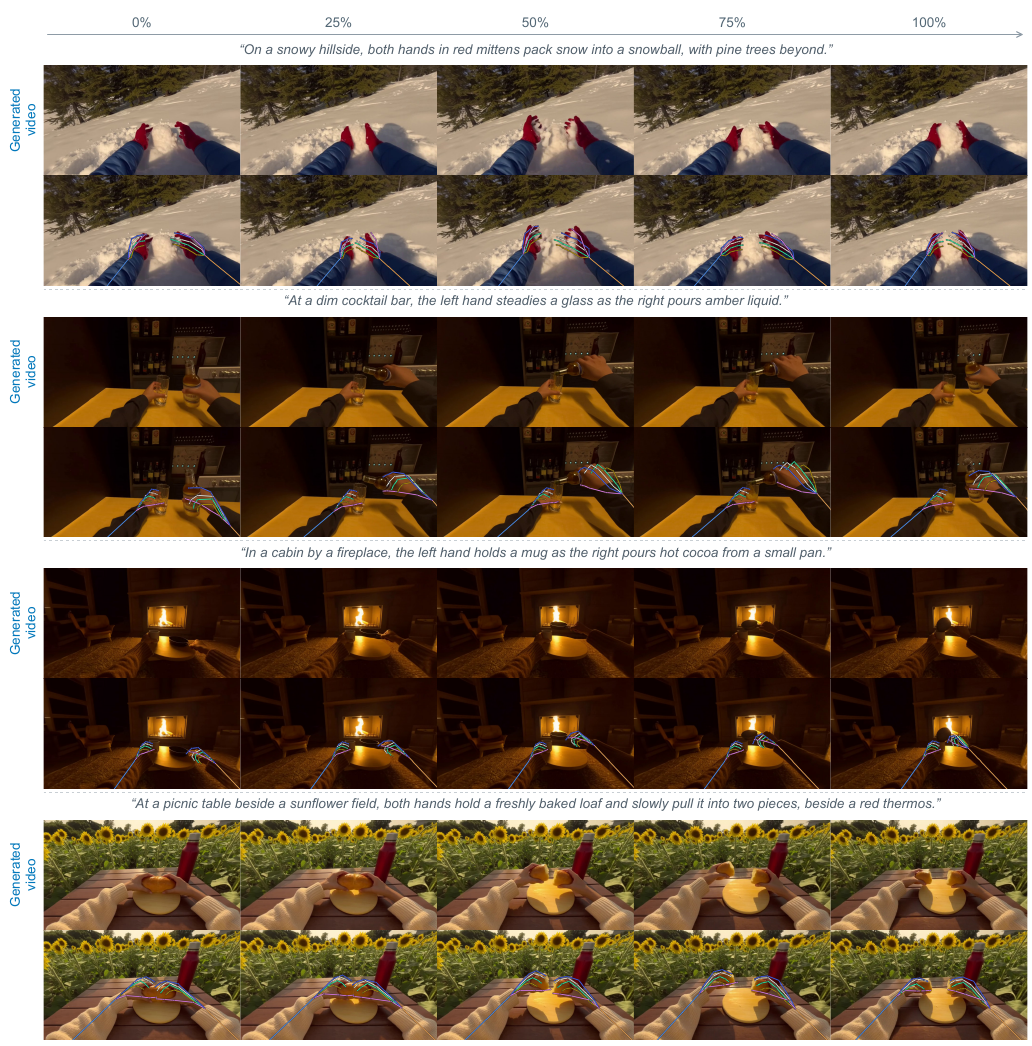}
  \caption{Additional text-only joint action--video generation.
  Across diverse scenes, generated actions remain well aligned with the overall hand-motion trends in the generated videos.
  Each example pairs generated video with decoded action overlays at frames 0, 10, 20, 30, and 40.}
  \label{fig:text-to-action-video-supp-1}
\end{figure*}

\begin{figure*}[t]
  \centering
  \includegraphics[width=\textwidth,height=0.93\textheight,keepaspectratio]{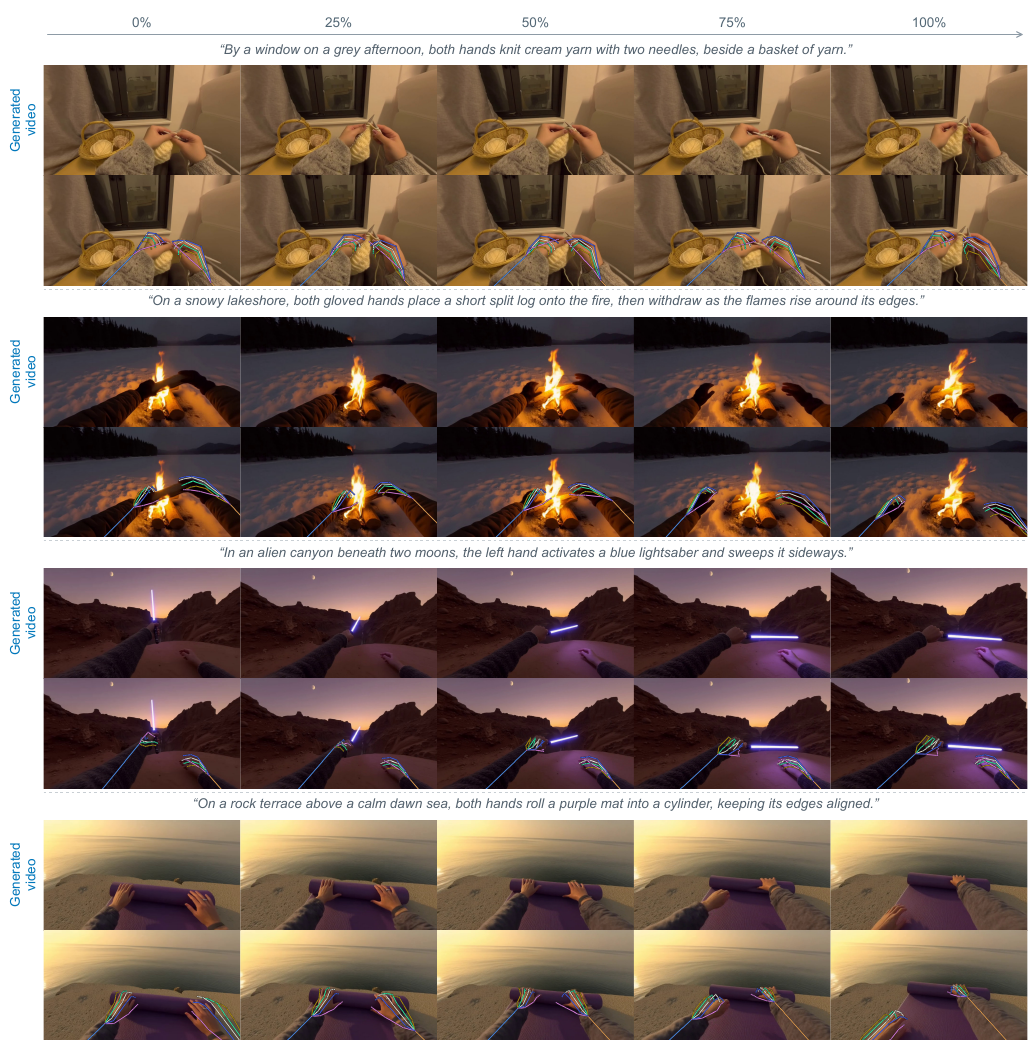}
  \caption{Further examples of text-only joint action--video generation.}
  \label{fig:text-to-action-video-supp-2}
\end{figure*}

\paragraph{Text-only joint action--video generation.}
\Cref{fig:text-to-action-video-supp-1,fig:text-to-action-video-supp-2} show synchronized video and motion generated from text alone across diverse scenes and interactions.

\subsection{Long-Sequence Generation}

\begin{figure*}[t]
  \centering
  \includegraphics[width=\textwidth]{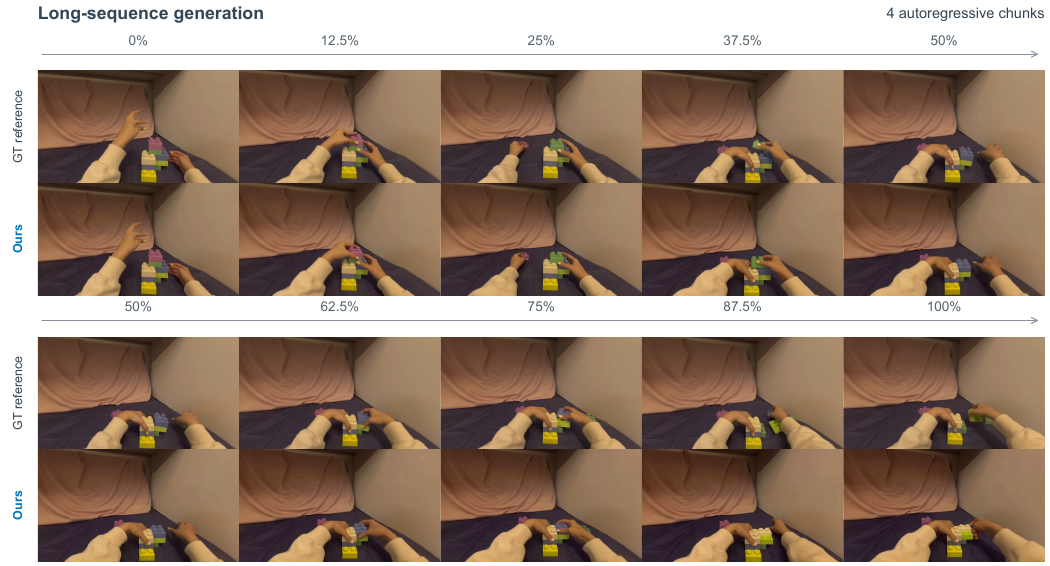}
  \caption{\textbf{Long-sequence generation.} \paper maintains physically coherent interactions over multiple autoregressive chunks.
  Four chunks are generated by conditioning each continuation on the preceding generated final frame.
  Alternating rows compare ground truth and generated video over the first and second halves of the sequence.}
  \label{fig:long-sequence-supp}
\end{figure*}

\Cref{fig:long-sequence-supp} demonstrates action-conditioned generation beyond the 41-frame training window, maintaining physically coherent object interactions over the extended sequence.

\subsection{Generalization to Edited Appearances and Environments}

\begin{figure*}[t]
  \centering
  \includegraphics[width=\textwidth,height=0.90\textheight,keepaspectratio]{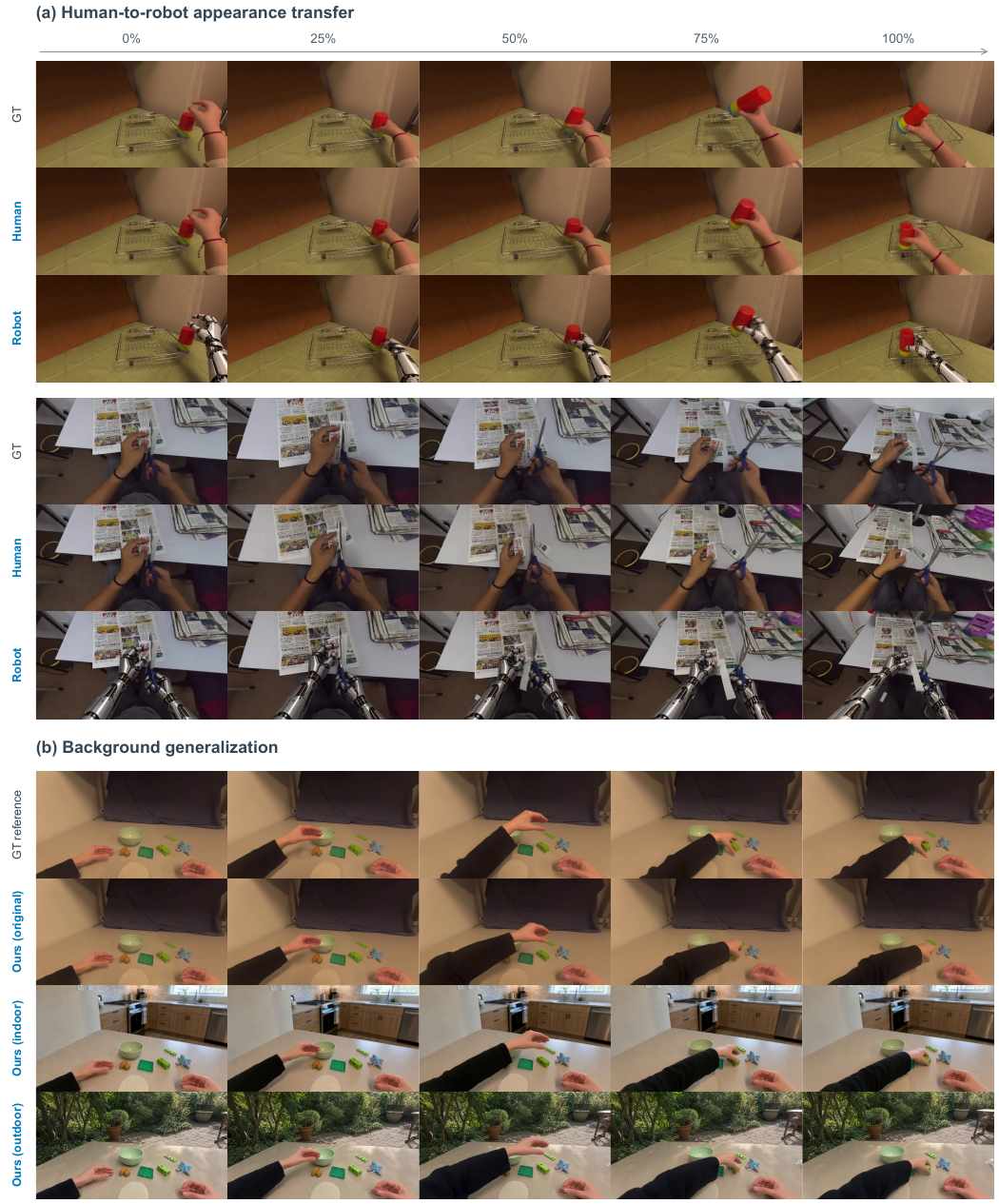}
  \caption{\textbf{Appearance and background variation.} \paper generalizes to changes in appearance and environment, opening up further applications of action-conditioned video generation. (a) Two human-to-robot appearance-transfer examples. (b) Ground truth and generated videos in the original scene and edited indoor and outdoor settings.}
  \label{fig:appearance-transfer-supp}
\end{figure*}

We evaluate action control under changes in hand appearance and scene background (\cref{fig:appearance-transfer-supp}).
With the first frame edited to show robotic hands or different indoor and outdoor settings, \paper continues to generate videos consistent with the prescribed action.

\clearpage
\clearpage

\end{document}